\documentclass[conference]{IEEEtran}
\ifCLASSINFOpdf
\else
\fi

\usepackage{url} %to enable url % by zkh
\usepackage{flushend} %to balance the last page
\usepackage{graphicx} % to enable jpg
\usepackage{subcaption} % to enable subfigure
\usepackage{epstopdf} % to enable eps
\usepackage{amsmath} % to enable equation*
\usepackage{amssymb} % diff
\usepackage{mathrsfs}
\usepackage[ruled,linesnumbered]{algorithm2e}
\usepackage{caption} % to change the property of caption
\usepackage{tikz} % to draw vector graphics
\usetikzlibrary{snakes}

\usepackage{soul}
\usepackage{times}
\usepackage{endnotes}
\usepackage{balance}
\usepackage{amsfonts}
\usepackage{epsfig}
\usepackage{amscd}
\usepackage{array}
\usepackage{graphics}
\usepackage{multirow}
\usepackage{breakurl}
\usepackage[T1]{fontenc}
\usepackage{CJKutf8}
\usepackage{tabularx}
\usepackage{enumerate}
\usepackage{caption}	%font for captions

\newcommand{\ignore}[1]{}
\newcommand{\revised}[1]{}

\newcommand\malurl[1]{\href{notalink}{\nolinkurl{#1}}}
\DeclareMathOperator*{\argmax}{argmax}

\begin{document}

\title{\Large \bf Query-Free Attacks on Industry-Grade Face Recognition Systems
              under Resource Constraints}
\author{\IEEEauthorblockN{Di Tang}
\IEEEauthorblockA{Chinese University of Hong Kong \\
td016@ie.cuhk.edu.hk}
\and
\IEEEauthorblockN{XiaoFeng Wang}
\IEEEauthorblockA{Indiana University\\
xw7@indiana.edu}
\and
\IEEEauthorblockN{Kehuan Zhang}
\IEEEauthorblockA{Chinese University of Hong Kong\\
khzhang@ie.cuhk.edu.hk}}
\maketitle

% Use the following at camera-ready time to suppress page numbers.
% Comment it out when you first submit the paper for review.
\thispagestyle{empty}

\begin{abstract}

To launch black-box attacks against a Deep Neural Network (DNN) based Face Recognition (FR) system, one needs to build \textit{substitute} models to simulate the target model, so the adversarial examples discovered from substitute models could also mislead the target model. Such \textit{transferability} is achieved in recent studies through querying the target model to obtain data for training the substitute models. A real-world target, likes the FR system of law enforcement, however, is less accessible to the adversary. To attack such a system, a substitute model with similar quality as the target model is needed to identify their common defects. This is hard since the adversary often does not have the enough resources to train such a powerful model (hundreds of millions of images and rooms of GPUs are needed to train a commercial FR system).

We found in our research, however, that a resource-constrained adversary could still effectively approximate the target model's capability to recognize \textit{specific} individuals, by training \textit{biased} substitute models on additional images of those victims whose identities the attacker want to cover or impersonate. This is made possible by a new property we discovered, called \textit{Nearly Local Linearity} (NLL), which models the observation that an ideal DNN model produces the image representations (embeddings) whose distances among themselves truthfully describe the human perception of the differences among the input images. By simulating this property around the victim's images, we significantly improve the transferability of black-box impersonation attacks by nearly 50\%. Particularly, we successfully attacked a commercial system trained over 20 million images, using 4 million images and 1/5 of the training time but achieving 62\% transferability in an impersonation attack and 89\% in a dodging attack.

\end{abstract}

\section{Introduction}
\label{sec:introduction}
With its commercial success, Deep Learning (DL) based Face Recognition (FR) is haunted by the security risks posed by the adversary who have already been adaptive to the AI innovation. Prior research shows that \textit{adversarial examples} can be found to mislead even the state-of-the-art recognition algorithms~\cite{goodfellow2015explaining, huang2017adversarial, szegedy2013intriguing, xu2017feature}, causing them to misclassify these examples. More specifically, such adversarial examples are images derived from adding perturbation on normal images, for the purpose of inducing classification errors while maintaining the level of changes low so they can appear less distinguishable from the original images by humans. Indeed, a recently approach~\cite{carlini2017towards} alters merely 16 pixels to ensure misclassification on 32$\times$32 images.

\vspace{5pt}\noindent\textbf{Attacking a straw-man}. On the other hand, such adversarial learning risks need to be put into perspective. Still we are less clear how \textit{realistic} the discovered threats could be, given that most of them are reliant on a \textit{white-box} assumption about the \textit{target} (the FR system they aim at), that is, the availability of full information about the target's parameters. In practice, however, an industry-grade system's parameters are often commercial secret and cannot be easily acquired by unauthorized parties.

A more realistic way to understand a DL system's security properties is the \textit{black-box} approach, in which the adversary \textit{queries} the target, utilizes the features inferred through the queries to learn a \textit{substitute} model and then searches for the adversarial examples that also work on the target model. Such an approach is based upon \textit{transferability} of adversarial examples across different models~\cite{liu2016delving}: some examples mislabeled by one DL model are also found to be misclassified by another. A direct attempt to transfer adversarial examples through an ensemble learning~\cite{liu2016delving} was found to be less effective. 
To ensure a high transferability, more recent approaches aggressively \textit{query the target} to obtain adequate input-output samples for accurately simulating the target model.  As a prominent example, a recent black-box attack needs to interact with the target for at least 1,000 times~\cite{papernot2017practical}.

With all the progresses being made, a big gap still exists between hypothetic attacks proposed and credible threats with practical impacts. Particularly, querying security-critical FR systems is often expensive or even infeasible in practice: e.g., an FR ATM can immediately alert a card holder to a potential fraud once an impersonation attempt fails, making further probes less likely to continue. Another problem of prior transferability studies is the simple dataset used to train their models, e.g., the MNIST database~\cite{lecun1998gradient} includes only tens of thousands of images for recognizing ten handwritten digits. A real-world FR system, however, is typically trained over tens or even hundreds of millions of images for identifying millions of identities. Less clear is whether what is learned from such small-scale studies over toy examples is indeed applicable to real FR systems.

\vspace{5pt}\noindent\textbf{Cross-class transferability}. To better understand the security guarantee of real-world FR systems, we revisited transferability in our research, assuming that the adversary \textit{cannot} get any feedback from the target model and has limited resources. In our study, we trained multiple common deep neural networks (DNN), including VGG, GoogLeNet and ResNet, and evaluated the transferability of adversarial examples across these models using the standard ensemble learning based attack reported in the prior research~\cite{liu2016delving}, under various settings (shadower substitute networks, different structures and fewer data) to simulate a resource-constrained attacker. This research sheds new light on transferability: e.g., for ResNet, the transferability from a 50-layer substitute to a 101-layer target is about 16.8\%, compared with 24.6\% between the 101-layer substitute and the same target, in an impersonation attack. More interesting is the significant impact of \textit{training data sizes}: the transferability has dropped from 24.6\% in an impersonation attack to 14.5\% when the substitute was trained on a dataset one order of magnitude smaller than that of the target, and further to 7.1\% for a training set two orders of magnitude smaller. Intuitively, substitutes learned with fewer data or a shallower model would have a looser boundary and thus is less likely to ensure a misclassification on the target (which is better trained with more data). Overall, we only witnessed a limited success on transferability, particularly when it comes to the impersonation attack: about 20\% under different settings.

\vspace{5pt}\noindent\textbf{Our work}. To attack an industry-grade FR system without querying it, a set of high-quality substitute models need to be built to find \textit{common} defects of the DNN models similar to or even better trained than the target model. However, constructing such substitute models is hard, particularly with limited resource. In our research, we studied what the adversary could do to narrow this gap and enhance his odds of success.  A unique observation we have is that even though the target model generally has a more precise decision boundary, the substitute model could still \textit{partially} approach this boundary in some regions: for example, a criminal may leveraging a large number of his own and his victim's photos to boost the substitute model's accuracy with regard to the identification of (just) these two individuals, for the purpose of finding the right makeup to cheat an FR ATM into authenticating him as the victim. This attack, which we call \textit{Asymmetric Cross-Class Image Transfer} or \textit{EXCIT}, is found to be completely feasible in our research, due to a new property called \textit{Nearly Local Linearity} (NLL) discovered in our study.

More specifically, under a well-trained DL model, the difference between a pair of images' \textit{representations} (i.e., the Cosine distance between the embedding vectors produced by the DL model) should be nearly linear to their similarity as seen by human eye: in other words, when these images become increasingly dissimilar, the difference between their representations grows large proportionally. This NLL property, as discovered in our research, can be approximated during the training process of our substitute models: using our synthesized additional images of victim and the attacker himself, we minimize the gap between the distances across different representations produced by the model and what are supposed to be according to NLL. We found that such a model can effectively simulate a better-trained target model's behaviors around the images of interest to the attackers. (For the sake of simplicity, in the following article, we call these images as the Points of Interesting or PoIs.)  

%we push the distances between the representations produced by the model to what they are expected according to NLL.

In our research, we implemented EXCIT and, first, evaluated it under the settings of our transferability study. We observed that the new technique vastly enhanced the effectiveness of the attacks, particularly for impersonations, from 20\% (based upon the prior attack~\cite{liu2016delving}) to 50\%, even when the adversary only used 10\% of the training data and half of the layers (thus saving the training time by 5 orders of magnitude). Further we ran this approach against industry-grade systems including ColorReco, Facevis, Face++ and SenseTime (the SenseTime system trained over tens of millions of photos). Using 4 million images collected from the web (the largest scale for this type of research), EXCIT was found to significantly elevate the chance of successful cross-model attacks compared with the naive query-free attack~\cite{liu2016delving}, from 11\% to 62\%, without any communication with the target model before the attack.  

%Further we demonstrate that the odds of transferability can be estimated for the examples discovered, and the overall cost of the attack is reasonable: one can readily obtain the resources and data for the attack (4 million photos from the Internet and 10000 dollars' machine, from Amazon).

\vspace{5pt}\noindent\textbf{Contributions}. The contributions of the paper are outlined as follows:

\vspace{3pt}\noindent$\bullet$\textit{ The NLL property and understanding of transferability}.  Our large-scale study reveals a fact that the training data size can have on the successful transferring of an adversarial instance from one model to another. More importantly, we discovered the nearly linear relation between input images and their representations (in terms of their differences) under an ideal model, which enables our query-free attack and might lead to better understanding of the fundamental defects in DL models. 

%We report the first study on the transferability between DNN models over a training dataset of a large scale. This study reveals the significant impact the training data size has on the success of transferring an adversarial instance from one model to another, which is not known before and important to understanding the security risks coming with transferability.

%We report the first study on the transferability between DNN models over a training dataset of a realistic scale, including 4 million images. This study reveals the significant impact the training data size has on the success of transferring an adversarial instance from one model to another, which is not known before and important to understanding the security risks coming with transferability.

\vspace{3pt}\noindent$\bullet$\textit{ New techniques for query-free attacks}. Based upon the new discovery, we designed a new attack technique that finds adversarial examples against a well-trained target model \textit{without querying the target and using limited resources}. At the center of the technique is leverage of additional synthesized images of victim and attacker and the NLL property to train substitute models that are capable of simulating the target model around PoIs, even when the adversary only possess a small amount of training data and much less computing resources. This makes an important step toward understanding the realistic threat of adversarial learning.  

%\vspace{3pt}\noindent$\bullet$\textit{ New techniques for query-free attacks}. Based upon the new discovery, we designed the first attack technique that finds adversarial instances against  \textit{off-line} from a substitute model for  attacking an \textit{online} model \textit{without out querying the target}.  At the center of the technique is our finding of the NLL property, which allows us to simulate the behavior of a better-trained model around PoIs, even when the adversary only possess a relatively small amount of training data and much less computing resources. This makes an important step toward understanding the realistic threat of adversarial learning.  
    
 \vspace{3pt}\noindent$\bullet$\textit{ Implementation and evaluation}.  We implemented the technique and evaluated it over industry FR systems.

%\vspace{5pt}\noindent\textbf{Roadmap}. The rest of the paper is organized as follows. Section~\ref{sec:background} introduces the background of our work and describes the adversary model and other assumptions. Section~\ref{sec:work} is the design of our protocol. Section~\ref{sec:securityAnalysis} describes security analysis; Section~\ref{sec:evaluation} elaborates the test settings and evaluation results.  Section~\ref{sec:relatedWork} compares our approach with related work. Section~\ref{sec:discussion} discusses the limitations of our technique and envisions the possible future research on mitigating the threats and Section~\ref{sec:conclusion} that concludes the paper.

%\item \textbf{Nearly Local Linearity.} We find that, in face identification field, a powerful DL model has significant NLL that is one kind of transferability. And we systematically demonstrate that this finding is irrelevant to the choosing of model structure.  
%\item \textbf{A practical black-box attack.} Based on NLL, we develop a black-box attack leveraging a bunch of pre-trained substitute models. This attack is query-free and far more practical than previous works.
%\item \textbf{Evaluation.} We 

\section{Background}
\label{sec:background}

\subsection{Deep Learning and Face Recognition}
\label{subsec:DNN}

\noindent\textbf{Deep Neural Network}. Deep Neural Network (DNN) is a function that projects the input domain onto an output domain for classification and other purposes.
Following the prior research~\cite{papernot2016effectiveness}, we formulate the DNN for image processing as below:
\begin{equation}
\notag
F(x) = softmax(Z(x)) = y
\end{equation}
where $x$ is the image serving as the input to the DNN, $y$ is its output, typically a vector of probabilities for the image to be in different classes, and  $Z(x)$ is the ``logits'',  the output of the layer right before the ``softmax'' layer, and, in other words, $Z(x)$ is the unscaled probability vector serving as the inputs to the ``softmax'' layer.

In all our concerned DNN structures (VGG, GoogLeNet and ResNet), the $Z(x)$ can be further decomposed as:
\begin{equation}
\notag
Z(x) = T \circ R(x)
\end{equation}
where $R(x)$ is the feature vector of the input $x$ extracted by our DNN model. Specifically, the DNN projects $x$ onto a feature space and $R(x)$ is the representation of the $x$ in that space. $T(\cdot)$ is the classification function that transforms $R(x)$ into ``logits''. Usually, $T(\cdot)$ is a linear mapping function and has the form of
\begin{equation}
\notag
T(R(x)) = W_C R(x) + B_C
\end{equation}
where $W_C$ is the weight matrix and $B_C$ is the bias vector.

A well-trained DNN is characterized by its capability to generate similar representations for similar inputs. This avoids the pitfall when two similar inputs actually are mapped to two very different representations and as a result, are assigned into two different classes. Note that in our research, the similarity between two representations is measured by the cosine distance between them.

\vspace{5pt}\noindent\textbf{Face recognition systems}. Since the introduction of deep Convolutional Neural Networks (CNN)~\cite{krizhevsky2012imagenet}, FR technologies have been evolving rapidly. As a prominent example, DeepFace~\cite{taigman2014deepface} close the gap between the recognition capabilities of human beings and machines. Further, DeepID3~\cite{sun2015deepid3} attained a 99.53\% accuracy on the LFW dataset~\cite{huang2007labeled} that exceeds the human performance, 99.2\%. More recently, FaceNet~\cite{schroff2015facenet} exploited a deep architecture to achieve a 99.63\% accuracy on the same dataset.

More generically in the image processing area, three DNN models have been extensively used. VGG-16~\cite{simonyan2014very} running 16 cascaded convolution layers was reported to achieve state-of-the-art recognition results in the ImageNet Large-Scale Visual Recognition Challenge 2014~\cite{ILSVRC15} (ILSVRC-2014), together with GoogLeNet~\cite{szegedy2015going}, which involves 22 layers and Inception architectures invented by Google for combining information from multi-views.  Empowered by the pervasiveness of GPU and Batch Normalization technologies~\cite{ioffe2015batch}, ResNet-152~\cite{he2016deep} winning the ILSVRC-2015 classification task is armed with 152 layers and capable of transferring shadow features to deep layers.

\subsection{Adversarial Learning}
\label{subsec:AL}

The potential of deploying DNN to real-world systems (e.g., self-driving cars) faces the security challenges of \textit{adversarial learning}, an attack that manipulates the inputs to a DNN to cause misclassification. This attack was first discussed by Szegedy et al.~\cite{szegedy2013intriguing}, who pointed out the existence of \textit{adversarial examples}, i.e., perturbed input $x'$ is similar to the original input $x$ but misclassified by the DNN into a different category. Such attacks can be targeted or not. In the non-targeted case, the attacker seeks adversarial examples that are misclassified into \textit{any} categories except the one they belong to. For instance, the adversary wants to fool the face recognition system by slightly changing his face and making it is misclassfied as other people. Formally speaking, the attacker changes facial appearance from $x$ to $x'$, and causes the missed classification result: $\argmax_i F(x')_i \neq \argmax_i F(x)_i$.  Here, the DNN output $F(x)$ is a vector that describes the probabilities for the input $x$ belonging to different individuals. In a targeted attack, the adversary intends to impersonate a given individual $t$, by seeking a makeup $x'$ causing $\argmax_i F(x')_i = t$. As we focus on FR problems, we will use dodging attacks to represent non-targeted attacks and impersonation attacks to represent targeted attacks.

\vspace{5pt}\noindent\textbf{Attack methods}. To find adversarial examples, people need to define the similarity between two images (the inputs), $x$ and $x'$, based upon a distance metric. Prior research on adversarial learning uses the $L_p$ distance, with $p$ being 0, 2 or $\infty$:
\begin{equation}
\notag
\|x-x'\|_p = (\sum_{i=1}^{n} |x_i-x'_i|^p)^{\frac{1}{p}}.
 \vspace{-0.1in}
\end{equation}
Here $x_i-x'_i$ is the subtraction between the $i$-th pixel of two input images. 

Minimizing the $L_0$ distance, we can get $x'$ with the smallest number of pixels differing from those on the original input $x$. The \textit{Jacobian-based Saliency Map} (JSMA)~\cite{papernot2016limitations} is an attack optimized under the $L_0$ distance. It iteratively picks pixels that have the most impact on the results and modifies them, until either a given threshold (an upper bound for the number of pixels) is reached or an adversarial example is found.

Minimizing the $L_2$ distance, we can obtain $x'$ that has the least modification, in terms of Euclidean distance, across all pixels on $x$ and $x'$. The first attempt using this distance is L-BFGS~\cite{szegedy2013intriguing} that minimizes the $L_2$ distance under the box-constraint, i.e., $x' \in [0,1]^n$, where $n$ is the number of pixels. It exploited the classical gradient descend method to find the optimal solution with a pre-defined learning rate $lr$: 
\begin{equation}
x' = x + lr \cdot \bigtriangledown_x F(x),
\label{eq:att_obj}
\end{equation}

Minimizing $L_{\infty}$ distance, we can find $x'$ with the smallest maximum-changes to the pixels. Under this distance, the optimization algorithm seeks a region of pixels with similar intensities  to modify. An example of the prior attack is \textit{Fast Gradient Sign Method} (FGSM)~\cite{goodfellow2015explaining}, which iteratively updates $x'$ to produce an adversarial example by stepping away a small stride along with the direction of $\bigtriangledown_x F(x')$. 

In our research, we chose $L_2$ distance, because it is a continuous metric and also sensitive to the change that happens to any pixels in the input images. By comparison, $L_0$ (number of different pixels) is not continuous and $L_{\infty}$ (maximum pixel difference) might not capture a small modification on the input.

%In our research, we chose $L_2$ distance, because it can provide a continuous and global vision to observe the procedure changing the input $x$ from $x^{(1)}$ to $x^{(2)}$. As the opposite, $L_0$ is not continuous and $L_{\infty}$ is not global.

\vspace{5pt}\noindent\textbf{Transferability}. As mentioned earlier, transferability is the key to practical adversarial learning, when the adversary cannot directly access the internal parameters of the target model. Prior research~\cite{liu2016delving} demonstrates that around 20\% adversarial examples discovered from one of the three models (ResNet-152, VGG-16 and GoogLeNet) are also misclassified by other two models under a dodging attack. A more recent study~\cite{papernot2016transferability} further shows that transferability can happen even across different machine learning techniques: DNN, Logistic Regression (LR), Support Vector Machine (SVM) and Nearest Neighbors (kNN). Particularly, more than 60\% of adversarial examples discovered in LR or SVM were found to be still effective on the other model. When it comes to the impersonation attack, also 20\% adversarial examples were reported to work across different DNN models~\cite{liu2016delving}. These examples were found using an ensemble-based approach that descends along the summation of the gradients of several models.

A primary limitation of these prior studies is that they are all based upon ``less-categories'' datasets, such as ILSVRC-2012 including 1000 categories. Compared with the industry-grade FR systems such as Facevisa, ColorReco, which are trained to classify tens or even hundreds of thousands of identities, what has been learned from these studies can be less conclusive. Also importantly, the prior research either considers that the substitute models are built upon similar or even identical datasets as the target, or at the very least, assumes that the adversary is capable of continuously querying the target model to collect data (query results) for training the substitute models. As discussed earlier, in many cases, these assumptions are still a far from reality. Our research instead looked into the transferability over a large dataset, when the adversary cannot query the target and has limited resource to train his substitute models.

\subsection{Threat Model}
\label{subsec:threat}

We consider an adversary that intends to perform a dodging attack or a impersonation attack on a target FR model that he \textit{cannot} query. The adversary does not have access to the internal parameters of the target but has limited information about its architecture (e.g., ResNet, VGG or GoogLeNet) and its depth (e.g., about 100 layers for ResNet, though the precise number of layers is still unknown to him).  All such information about a commercial system is often made available through various public sources, such as research papers (e.g., the design of Face++ was described in the paper~\cite{fan2014learning}), technical reports and other online documents.

The target model studied in our research is assumed to be trained over a large amount of data, tens or even hundreds of millions of images, as those commercial FR systems are. On the other hand, the adversary does not have that level of resources, though we do assume that he can still acquire millions of images publicly available online, as we did in the study.  Further, the adversary can obtain thousands of images of himself and the victim he want to impersonate, and also sufficient resources from the cloud to train the substitute model over the data. We believe that these assumptions are all realistic, as demonstrated in our research: particularly, all the computing power required for training our attack model can be purchased from Amazon at an approximate cost of 10,000 dollars. Specially, if the adversary can not obtain sufficient images of the victim from the Internet, they can follow the victim and record videos to get enough images that are taken in various scenarios and from different angles.

\section{Understanding Transferability across Asymmetric Models}
\label{sec:unsderstand}

To understand whether a DNN model is vulnerable to query-free attacks, we need to find out the challenges in simulating the target model's behaviors, under the limited resources and information.  For this purpose, we conducted the largest study on transferability, using a dataset with 4 million images.  Our research reveals the importance of training data size to a successful cross-model attack.

\subsection{Settings}

Our study utilized \textit{MegaFace Challenge 2}~\cite{nech2017level}, a dataset including 672K identities and their above 4 million photos, and \textit{Caffe}~\cite{jia2014caffe}, an open-source deep learning framework, to train FR DNN models in our experiments. All such experiments were conducted on a 8-GPUs server with each GPU armed with 12GB memory.

In our studies, we assume our target model is $F^*(\cdot)$ and it outputs a vector $F^*(x)$, for a input image $x$. Our study covers both dodging attacks and the impersonation attacks. The criteria for a successful dodging attack is:
\begin{equation}
\notag
\begin{array}{c@{\quad}l@{\quad}l}
& \mbox{find } & x'  \\
s.t. & \|x-x'\|_2 &\leq \theta \\
& \argmax_i F^*(x')_i &\neq o
\end{array}
\end{equation}
where $o$ is the owner of $x$, $o = \argmax_i F^*(x)_i$.
The criteria for a successful impersonation attack is:
\begin{equation}
\notag
\begin{array}{c@{\quad}l@{\quad}l}
& \mbox{find } & x'  \\
s.t. & \|x-x'\|_2 &\leq \theta \\
& F^*(x')_t &> 0.5
\end{array}
\end{equation}
where $t$ is the victim to be impersonated and we ensure $t \neq o$. For the better understanding, during later paragraphs, we may use ``subject'' to indicate the one whose identity is $o$ and ``victim'' to indicate the one whose identity is $t$. 

In following studies, we will use 4 levels of training dataset to train our models. For the sake of clarity, we list them in Table~\ref{tb:data_types}.  

\begin{table}[tbh]
  \renewcommand{\arraystretch}{1.5}
  \centering
    \caption{Different levels of dataset.}
	\begin{tabular}{|c|c|c|}
    \hline
    Short name & Images & Identities\\
    \hline
    L1 & $6 \times 10^4$ & $10^4$\\
    \hline
    L2 & $6 \times  10^5$ & $10^5$\\
    \hline
    L3 & $19 \times  10^5$ & $3 \times  10^5$\\
    \hline
    L4 & $40 \times  10^5$ & $6.4 \times  10^5$\\
    \hline
	\end{tabular}
	\label{tb:data_types}
\end{table}
Besides, we say the transferability between two models is $50\%$ which means there are $50\%$ adversarial examples found on one model that can successfully fool both two models. Sometimes we may use ``success rate'' to replace transferability. 

In this study, we use C\&W approach~\cite{carlini2017towards}, the best method as we know, to find adversarial examples. It optimizes the following objective function:
\begin{equation}
\notag
\begin{array}{c@{\quad}l}
\mbox{minimize} & \displaystyle{ \frac{1}{2} \|tanh(w)-x\|^2_2 + c \cdot f(tanh(w)) }.
\end{array}
\end{equation}
For the dodging attack, $f$ is defined as
\begin{equation}
\notag
f(x') = max(Z(x')_o - max\{ Z(x')_i : i \neq o\}, -\kappa).
\end{equation}
For the impersonation attack, $f$ becomes
\begin{equation}
\notag
f(x') = max(max\{ Z(x')_i : i \neq t\} - Z(x')_t, -\kappa).
\end{equation}
Here, the adversarial example we found is $x'=\tanh(w^*)$, where $w^*$ is the optimal solution of above function. In that function, $c$ is a parameter that balances the importance of two components, the first component minimizing the $L_2$ distance between the adversarial example $x'$ and the original image $x$, the second component modeling the goal of this attack, either dodging or impersonation. Also, $\kappa$ is a threshold indicating when the attack goal is achieved. We set $c=20$, $\kappa=20$ for both the dodging attack and the impersonation attack in our experiments.

And we further improve the performance of above function by exploiting the standard ensemble-based approach. Specifically, we will use $K=4$ substitutes and assemble them to solve the following function:
\begin{equation}
\begin{array}{c@{\quad}l}
\mbox{minimize} & \displaystyle{ \frac{1}{2} \|tanh(w)-x\|^2_2 + c \cdot \sum_{k=1}^{K} f^{(k)}(tanh(w)) }.
\end{array}
\label{eq:ensemble}
\end{equation}
where $f^{(k)}(\cdot)$ is the $k$-th substitutes.

\subsection{Impacts of Structural Features}
\label{subsec:structure}

\noindent\textbf{Structures}. As mentioned earlier, to understand the impacts of DNN structures on transferability, we looked into the three most prominent structures: VGG\footnote{\tiny{\url{https://github.com/davidgengenbach/vgg-caffe}}}, GoogLeNet \footnote{\tiny{\url{https://github.com/BVLC/caffe/tree/master/models/bvlc-googlenet}}} and ResNet\footnote{\tiny{\url{https://github.com/KaimingHe/deep-residual-networks}}}. In this study, for every structure, we trained a target model on a $L2$ dataset and four substitutes on four $L2$ datasets. And, we ensure that these five training datasets are not overlap with each other except for the images of subjects and victims involved in attacks.

Over these models, we analyzed the transferability of the dodging and impersonation attacks, using an ensemble learning method that integrates adversary examples found in 4 substitutes trained independently to find images causing the target model to misclassify. More specifically, in dodging attacks, we built an attacking set containing 635 images from 100 identities, and in impersonation attacks, we use 600 image-pairs to construct the attacking set. Each image-pair contains two images from two different identities. To be noticed that the attacking set were inside the training sets of both the target model and substitute models.

The experiment results are presented in Table~\ref{tb:struct:dod_imp}. As we can see from the table, for dodging, we observed a transferability about 95\%, and for impersonation, it became 20\%. This finding is pretty much in line with what is reported in the prior research, indicating that transferring adversarial examples across different models are feasible, though less effective in the case of impersonation.

\begin{table*}[tbh]
  \renewcommand{\arraystretch}{1.5}
  \centering
    \caption{Transferability among different structures. The number in every cell is the success rate corresponds to that using four substitute models with row structure to attack the target model with the column structure.}

	\begin{tabular}{|c|c|c|c|c|c|c|}
	\hline
	  \multirow{2}*{}& \multicolumn{2}{|c|}{ResNet-101} & \multicolumn{2}{|c|}{GoogLeNet} & \multicolumn{2}{|c|}{VGG-16} \\
	\cline{2-7}
	  ~ & Dodging & Impersonation & Dodging & Impersonation & Dodging & Impersonation\\
    \hline
    ResNet-50 & 95.1\% & 16.8\% &  \multicolumn{2}{|c|}{-} & \multicolumn{2}{|c|}{-} \\
    \hline
    ResNet-65 & 95.6\% &  18.0\% &  \multicolumn{2}{|c|}{-} & \multicolumn{2}{|c|}{-} \\
    \hline
    ResNet-80 & 96.3\% &  23.2\% &  \multicolumn{2}{|c|}{-} & \multicolumn{2}{|c|}{-} \\
	\hline  
	ResNet-101 & 98\% & 24.6\% & 96.7\% & 20.2\% & 96.4\% & 20.3\%\\
	\hline
    GoogLeNet & 95\%  & 16.5\% &  97.8\% & 23.1\% & 94.6\% & 18.5\% \\
	\hline
	VGG-16 &  93.4\% & 17\% & 94.4\% & 18.1\% &  97.2\% & 22.3\% \\
	\hline
	\end{tabular}
	
	\label{tb:struct:dod_imp}
\end{table*}

\vspace{5pt}\noindent\textbf{Depths}. Further we looked into the impacts of depths on transferability.  For this purpose, we utilized ResNet, since the depth of its structure can be easily adjusted. More specifically, we built 4 ResNets with 50, 65, 80 and 101 cascaded convolutional layers respectively. The compositions of their structures are presented in Table~\ref{tb:struct:dod_imp}. Using these structures, we also trained models to repeat experiments that have been done to inspect the impact of different structures (four substitute models trained on four L2 training sets to attack the target model trained on L2 training set).

\begin{table}[tbh]
  \renewcommand{\arraystretch}{1.5}
  \centering
    \caption{Structures of different depths. ResNet structures can be divided into 5 stages, starting with a convolution layer, followed by four stages, each including a different number of ``bottleneck'' blocks.}

	\begin{tabular}{|c|c|c|c|c|}
	\hline
	 & ResNet-50 & ResNet-65 & ResNet-80 & ResNet-101\\
	\hline
	Stage 1 & 1 Conv & 1 Conv & 1 Conv & 1 Conv \\
	\hline
	Stage 2 & 3 Blocks & 3 Blocks & 3 Blocks & 3 Blocks \\
	\hline
	Stage 3 & 4 Blocks & 4 Blocks & 4 Blocks & 4 Blocks\\
	\hline
	Stage 4 & 5 Blocks & 10 Blocks & 15 Blocks & 22 Blocks \\
	\hline
	Stage 5 & 3 Blocks & 3 Blocks & 3 Blocks & 3 Blocks\\
	\hline
	\end{tabular}

	\label{tb:resnet_depth}
\end{table}

In this study, we ran those models as substitutes to attack the target (ResNet-101), both dodging and impersonation. As expected, the complexity of the network (its depth) indeed affects transferability: more layers make the DNN more capable and enhance transferability. Again, transferability tends to be low for the impersonation attack, around 16\% when using ResNet-50 to attack ResNet-101.

\subsection{Impacts of Data Size}
\label{subsec:size}

\noindent\textbf{Training size and transferability}. An important observation is that a real-world adversary typically cannot get as many photos as a large organization uses to train its industry-grade FR system. An important question we were asking is what impacts a relatively smaller dataset could have on the chance of a successful cross-model attack.  For this purpose, we trained VGG, GoogLeNet and ResNet-101 models on three levels of datasets: L1, L2 and L3. Again, all these individuals were randomly drawn from our dataset and we made sure that there was no overlap across substitutes' training set and target model's training set. In this study, we utilized substitute models of the same structure with the target model for both the dodging attack and impersonation attack. The target model is built upon a L3 training set. The results are presented in Table~\ref{tb:dataset_size}. As we can see here, training data size turns out to significantly affect transferability and such an influence is also consistent across different structures.  Compared with the structural impacts, transferability became lower when we reduced the data size from L3 to L2. Compared with the impact of depth, the attack was less likely to succeed when we downsized the size of training set (L3 to L2) than when we removed layers (from 101 to 50).

\begin{table*}[tbh]
\renewcommand{\arraystretch}{1.5}
  \centering
    \caption{Transferability among different levels of training set. The number in every cell is the success rate corresponds to that using four substitute models trained on row settings to attack the target model trained on column settings.}

	\begin{tabular}{|c|c|c|c|c|c|}
	\hline
	   \multicolumn{2}{|c|}{} & \multicolumn{2}{|c|}{L3 dataset} & \multicolumn{2}{|c|}{L2 dataset} \\
	\cline{3-6}
	  \multicolumn{2}{|c|}{} & Dodging & Impersonation & Dodging & Impersonation \\
    \hline
    \multirow{3}*{ResNet-101} & L3 dataset &98.8\% & 25.2\% & \multicolumn{2}{|c|}{-} \\
    \cline{2-6}
    ~ & L2 dataset & 81.3\% & 17.5\% &  \multicolumn{2}{|c|}{-} \\
    \cline{2-6}
    ~ & L1 dataset & 34.5\% & 7.1\% &  71.2\% & 14.5\% \\
    \hline
    \multirow{3}*{GoogLeNet} & L3 dataset & 97.3\% & 24.1\% & \multicolumn{2}{|c|}{-} \\
    \cline{2-6}
    ~ & L2 dataset & 79\% & 16.2\% &  \multicolumn{2}{|c|}{-} \\
    \cline{2-6}
    ~ & L1 dataset & 32\% & 5.1\% &  69.4\% & 13.3\% \\
    \hline
    \multirow{3}*{VGG-16} & L3 dataset & 97.5\% & 23.9\% & \multicolumn{2}{|c|}{-} \\
    \cline{2-6}
    ~ & L2 dataset & 77\% & 16.3\% &  \multicolumn{2}{|c|}{-} \\
    \cline{2-6}
    ~ & L1 dataset & 32.3\% & 6.3\% &  66.7\% & 12.9\% \\
    \hline
	\end{tabular}

	\label{tb:dataset_size}
\end{table*}

Further, we trained four substitute models on four L1 training sets and use these substitute models to attack a model that was trained on L2 dataset. This experiment's results are illustrated in the right column of Table~\ref{tb:dataset_size}.  An interesting observation is that the transferability from L1 models to L2 models actually is lower (14.5\%) than that for L2 to L3 (17.5\%), even though the difference in the training sets is even larger in the latter case ($13 \times 10^5$ images) than the former ($5.4 \times 10^5$ images). Intuitively, with the increase in training set, a substitute model becomes closer to a perfect model and adding more data then be less effective in improving the model's precision than the time when the model only learns from a small size of training set and therefore much less accurate. Further analysis of the observation leads to the conclusion that the enhancement of transferability will slow down when the data size goes up (see Appendix~\ref{app:transferability}).

\vspace{5pt}\noindent\textbf{Discussion}.  Our study shows that although both structural features and the size of training set affect transferability, apparently the impact of the latter is more prominent. In practice, the structural information of many commercial FR systems can often be found, from research articles, public paper and other sources. On the other hand, a deeper network with more data certainly need more computing resources to train.  For example, on our system with 8 GPUs, training a ResNet-101 model took 9 hours for a L1 training set, while only half of the time was needed for a 50-layer model over the same set. Most importantly, collecting a large number of high-quality images is often a challenge for the adversary: for example, SenseTime Ltd's model is reported to be built from above 20M images and the dataset of this scale could \textit{not} be found on the Internet, up to our knowledge. Therefore, we believe that whether transferability could be enhanced in the presence of a relatively small set of training data is critical question for assessing the practical impacts of adversarial learning on FR systems. 

Also to attack a real-world system without querying it, the adversary needs to estimate his chance of success based upon the features of a given adversarial example, for example, the percentage of the pixels modified. This can \textit{not} be easily done since the adversary does not have access to the target system and therefore cannot figure out the probability of success by testing his adversarial examples on the target. However, our study described above shows that the transferability between the substitute models and the target model can actually be gagged using the transferability between a target model learned from a smaller dataset and smaller substitute models.  This is because the probability of success in the latter case is expected to be higher than that in the former. A more formal analysis of this observation is presented in Appendix~\ref{app:transferability}.

\section{Query-Free Asymmetric Attack}
\label{sec:approach}

To enhance transferability, ideally we need to make the substitute model very similar to or even more accurate than the target model. Although this is nearly impossible for most real-world adversaries, given their limited resources, particularly a much smaller training set they are able to obtain, still something can be done to narrow the gap between the two models. A key observation is that a unique resource the adversary often has is abundant photos of the subject (often himself) in a dodging attack and also those of the victim in an impersonation attack. Leveraging such images, we could train a model \textit{biased} toward the subject or the subject and victim pair. Even though such a model may be overfitting and therefore its overall accuracy could may be below that of the target model, all we care about here is just the target model's behavior around the subject and/or the victim (PoIs), which we could potentially simulate in the substitute models using this resource (extra photos). 

However, effective use of such resource turns out to be challenging. Table~\ref{tb:naive_aug} shows the experimental results when we directly duplicate those photos of subjects and the victims to the same size with the original dataset (L2 level). And use the duplicated images and original images to enhance the transferability under the VGG, GoogLeNet and ResNet models in attacking the target models trained over L3 dataset. From the table, we do not see a significant improvement in the effectiveness of the attack, compared with those without such data augmentation. 

\begin{table}[tbh]
  \renewcommand{\arraystretch}{1.5}
  \centering
    \caption{Transferability on naively augmented training set. Use four L2 substitutes to attack L3 model. The number in the bracket is the original transferability copied from Table~\ref{tb:dataset_size}.}
	\begin{tabular}{|c|c|c|}
	\hline
	  & Dodging & Impersonation\\
    \hline
	ResNet-101 & 83\%(81.3\%) &  18.2\%(17.5\%)  \\
	\hline
    GoogLeNet & 80.2\%(79\%) &  15.8\%(16.2\%)  \\
	\hline
	VGG-16 & 77.3\%(77\%) &  16.8\%(16.3\%) \\
	\hline
	\end{tabular}
	
	\label{tb:naive_aug}

\end{table}

Intuitively, the subject and victim's photos alone are insufficient for simulating the \textit{relations} established by a better trained model between them and between the subject and other identities in the dataset. Such relations need to be built upon other images and the way a well-trained DNN maps the input to feature vectors. Following we show how such relations can be modeled using a new property discovered in our research, called \textit{Nearly Local Linearity} (NLL), the key technique behind our EXCIT attack, which helps improving the substitute model for simulating the target model's behaviors around the subject and victims, boosting the transferability from below 30\% (for impersonation) to above 60\% on commercial systems (Section~\ref{subsec:realsystem}).

\subsection{Nearly Local Linearity}
\label{subsec:nll}

\ignore{
Essentially, an adversarial example fools an imperfectly trained model by inducing its \textit{nonlinear} behavior, that is, a small perturbation causing a big representation change.
}

A well-trained model will produce more accurate results than the poor-trained model. However, this property is useless for obtaining better transferability. Thus, we need more delicate findings.

\noindent\textbf{Observation}. The key idea is that the representations produced by an ideal DNN model should be accurately model the human perceptions: when two images look very different, the distance between the representations should be large, and when the images appear to be similar, the distance should become small. So we built experiments to figure out how the representation of a well-trained DNN model changes during the procedure changing the input image from the subject to the victim. And, to measure the changing, right metrics need to be chosen. In our research, we found that $L_2$ (as used in the prior research~\cite{carlini2017towards}) and Cosine distances can serve these purposes.   

In our research, we trained three ResNet-101 models on three different datasets: L1, L2 and L4. From each dataset, we selected, uniformly at random, $10^4$ pairs of images $(x_a,x_b)$, with the images from two different identities (identity $a$ and identity $b$) in one pair. Then between each image pair $(x_a,x_b)$, we synthesized a series of 99 images by equidistant interpolation. Formally, the $k$-th image can be represented as: 
\begin{equation}
\notag
x^{(k)} = x_a + \frac{k}{100}(x_b-x_a)
\end{equation}
Specially, $x^{(0)} = a$ and $x^{(100)} = b$.

\begin{figure*}[ht]
	\centering
  \begin{subfigure}{0.32\textwidth}
		\includegraphics[width=\textwidth]{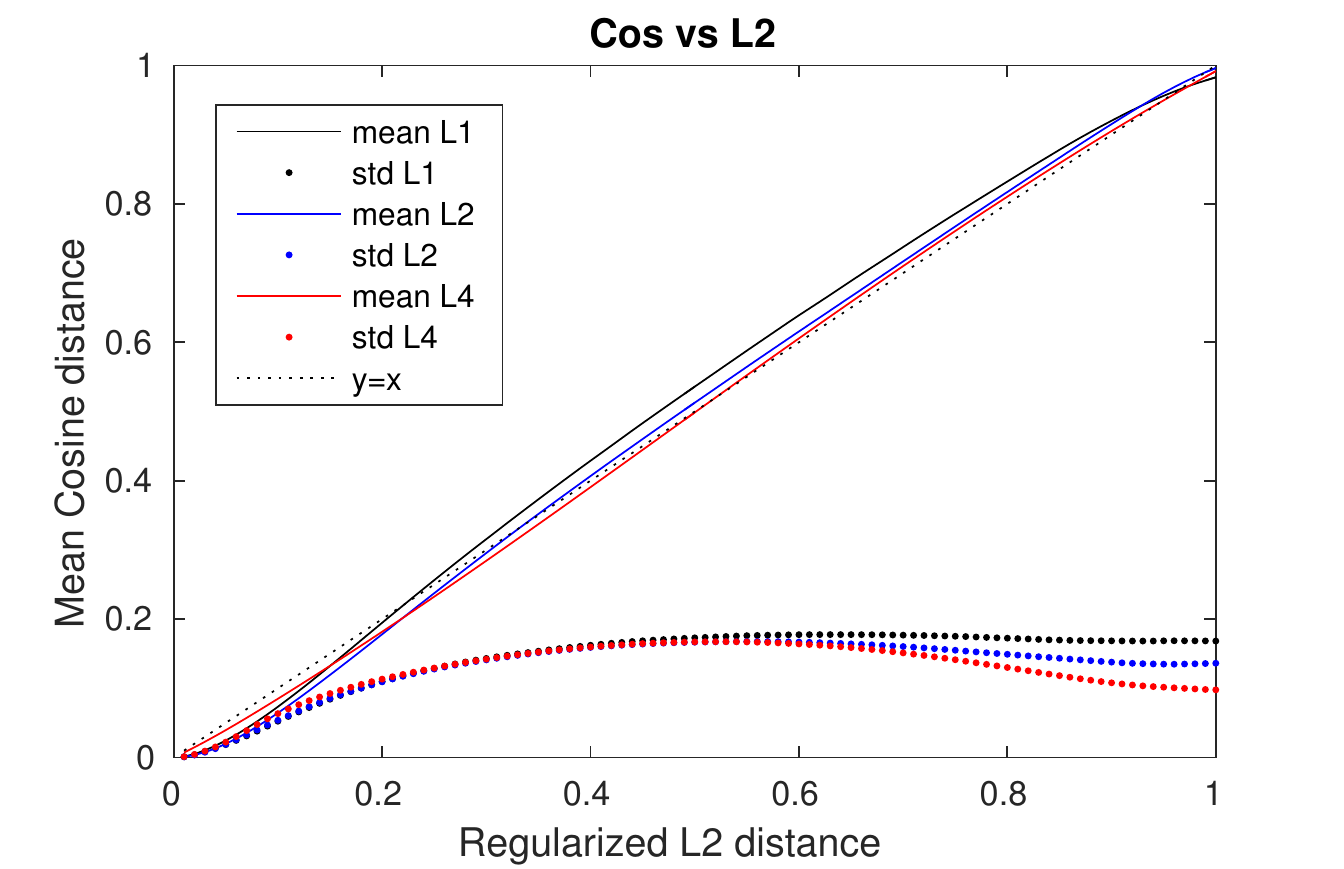}
        \caption{$\xi = 0.028, 0.018, 0.011$ for models of $L_1$, $L_2$ and $L_4$ respectively.}
   \end{subfigure}
   \begin{subfigure}{0.32\textwidth}
		\includegraphics[width=\textwidth]{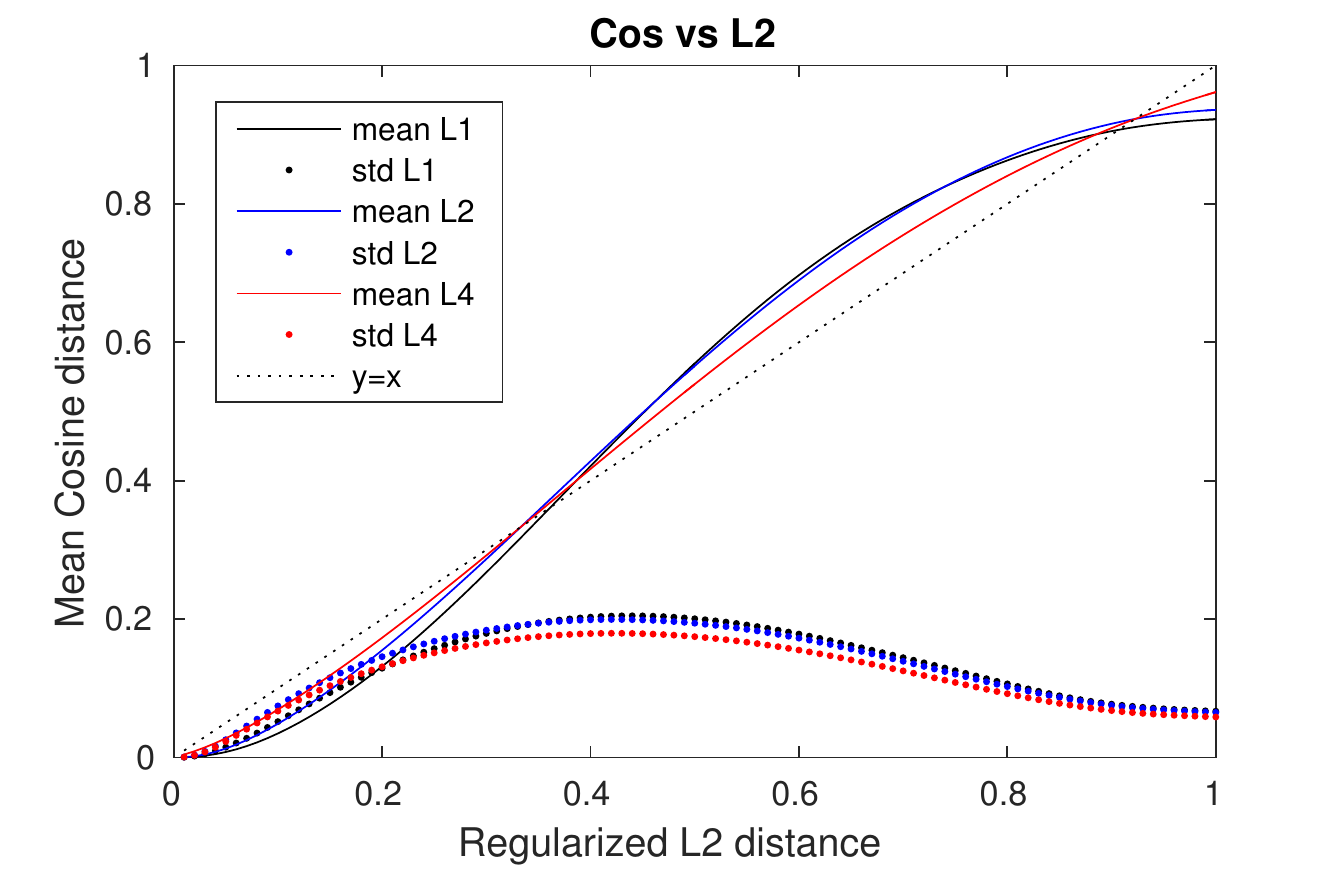}
        \caption{$\xi = 0.063, 0.047, 0.039$ for models of $L_1$, $L_2$ and $L_4$ respectively.}
  \end{subfigure}
  \begin{subfigure}{0.32\textwidth}
		\includegraphics[width=\textwidth]{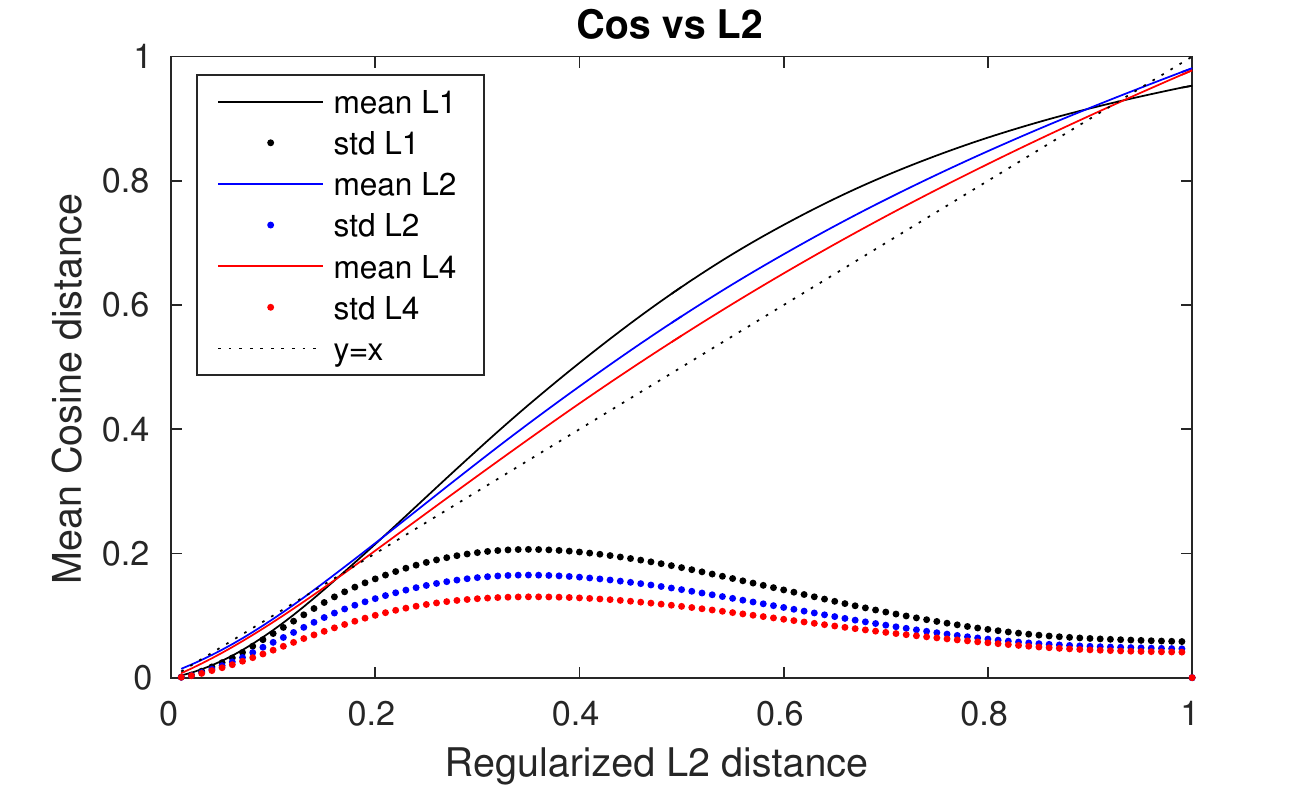}
        \caption{$\xi = 0.082, 0.053, 0.042$ for models of $L_1$, $L_2$ and $L_4$ respectively.}
  \end{subfigure}
	\caption{NLL property: From left to right, there are 3 figures showing the results of ResNet-101, VGG-16 and GoogLeNet models respectively.}
	\label{fig:rst_li_rsn}

\end{figure*}

Then, we ran all three models on these interpolated images to get their representations. Altogether, $10^6$ representations were produced from the $10^4$ image pairs. Further we calculated the mean for the Cosine distances between every $x^{(k)}$ and $x_a$ as:
\begin{equation}
\notag
\begin{array}{c@{\quad}l@{\quad}l}
& \bar{C}_a(k) = 10^{-4} \sum_{(x_a,x_b)}{C_a(k)} \\
\end{array}
\end{equation}
where 
\begin{equation}
\notag
\begin{array}{c@{\quad}l@{\quad}l}
& C_a(k) = 1-cos(R(x^{(k)}),R(x_a)) \\
\end{array}
\end{equation}
After that, we compare $\bar{C}_a(k)$ with the corresponding regularized $L_2$ distance $L_2(k) = \|x^{(k)}-x_a\|_2 / \|x_b-x_a\|_2$. The results are shown on Fig~\ref{fig:rst_li_rsn}. 

To more clearly demonstrate the different behaviors of different models, on Fig~\ref{fig:rst_li_rsn}, we also listed every model's $\xi$, the metric to measure the difference between the mean Cosine distance curve and the diagonal $y=x$:
\begin{equation}
\notag
\xi^2 = 10^{-2} \sum_{k=1}^{100}{(\bar{C}_a(k)-L_2(k))^2} 
\end{equation}

As we can see from the figures, the relation between the $L_2$ distance and the Cosine distance approaches linear with the increase of the training data size. Particularly, it becomes almost linear for the ResNet-101 model trained on L4 dataset (4 million images of $6.4 \times 10^5$ identities), with $\xi = 0.011$. Under the identical experimental settings, we observed the same $L_2$ and Cosine distance relation in the VGG-16 and GoogLeNet models. This indicates that the relations between the subject and interpolated images, and between the representations of them can be captured by this NLL property.

\vspace{5pt}\noindent\textbf{Concept}. 
Formally, we define the Nearly Local Linearity (NLL) as follows:
\begin{equation}
\begin{array}{c@{\quad}l}
1-\frac{R(x^{(\lambda)}) \cdot R(x_a)}{\|R(x^{(\lambda)})\|_2 \|R(x_a)\|_2} \approx \lambda = \frac{\|x^{(\lambda)} - x_a\|_2}{\|x_b - x_a\|_2}\\
\end{array}
\end{equation}
where
\begin{equation}
\notag
\begin{array}{c@{\quad}l}
x^{(\lambda)} = \lambda x_a + (1-\lambda) x_b, \lambda \in [0,1] \\
\end{array}
\end{equation}

\subsection{The EXCIT Attack}
\label{subsec:augment}

The discovery of the Nearly Local Linearity property in a well-trained DNN model enables us to train a substitute model by not only leveraging the extra images of subjects and victims but also integrate such images into the model by approximating the relations between synthesized images from those images and the existing images in the dataset. For this purpose, we need to build a set of ``transitional'' images as those interpolated images mentioned earlier, and redefine the optimization goals when training the substitute model to connect these images to others in an expected way. These are the key steps for our EXCIT attack, as elaborated below. 

\vspace{5pt}\noindent\textbf{Subject-oriented data augmentation}. To synthesize ``transitional'' images and enrich the training dataset, we designed a subject-oriented data augmentation algorithm (see Algorithm 1). This algorithm follows a key rule is to keep balance. Detailedly, the algorithm keeps balance in three levels: first, it keeps the total number of synthesized images being similar with the number of original images; second, it keeps the number of synthesized images between every image-pairs being the same; third, it uses the uniform distribution to control the generation of $\lambda$. Thus, adversaries can use limited images to synthesize as many as possible behaviors of a well-trained DNN model.

In our algorithm, again, $o$ is the subject and $t$ is the victim. For the sake of simplicity, we set $t=o$ to represent the dodging attack. Our algorithm takes the original dataset $\mathcal{D}$, a subject-victim pair $(o,t)$ and the number of images needed to be synthesized between one image-pair $m$ as its input and outputs the augmented dataset $\mathcal{D}_{aug}$. In the rest of the paper, we keep $m$ being 10.

\ignore{
For a dodging attack, $o$ and $t$ will be set to the same input so the algorithm will look for transitional images between all $o$'s images $\mathcal{A}$ and randomly-selected other images $\mathcal{B}$ in the dataset. Given the constraint of the total number of synthetic images $n$ and the requirement that for each image pair, at least 10 synthetic photos need to be injected, we have $|\mathcal{B}| \leq \frac{n}{10}/|\mathcal{A}|$. For each photo pair $(a, b)$ from $\mathcal{A} \times \mathcal{B}$, our approach randomly ($\lambda \sim U(0,1)$) interpolate $\frac{n}{|\mathcal{A}| |\mathcal{B}|}$ transitional images.
}

\begin{algorithm}[htb]
\label{alg:aug}
\caption{Subject-oriented data augmentation algorithm.}
  \KwIn{$\mathcal{D}$, $(o,t)$, $m$}
  \KwOut{$\mathcal{D}_{aug}$}
  $\mathcal{A} = \{x_o: x \in \mathcal{D}, argmax_i \ F(x)_i = o \} $\;
  $\mathcal{B} = \{x_t: x \in \mathcal{D}, argmax_i \ F(x)_i = t \} $\;
  $\mathcal{C} = \mathcal{D} - \mathcal{A} \bigcup \mathcal{B}$\;
  \If{$o=t$}{
    $n = 0$\;
  }
  \Else{
    $n = |\mathcal{D}| / m$\;
  }
  pairs = []\;
  \For{$i=1$ to $n$} {
    randomly select a $x_a$ from $\mathcal{A}$\;
    randomly select a $x_b$ from $\mathcal{B}$\;
    pairs.append(($x_a$,$x_b$)) \;
  }
  \For{$i=1$ to $|\mathcal{D}|-n$} {
    randomly select a $x_a$ from $\mathcal{A} \bigcup \mathcal{B}$\;
    randomly select a $x_b$ from $\mathcal{C}$\;
    pairs.append(($x_a$,$x_b$)) \;
  }
  \For{ ($x_a$,$x_b$) in pairs} {
    \For {i = 1 to m} {
      Sample $\lambda$ from $U(0,1)$\;
      $x^{(\lambda)} = \lambda x_a + (1-\lambda) x_b$\;
      $\mathcal{D}_{aug} = \mathcal{D}_{aug} \cup \{x^{(\lambda)}\}$\;
    }
  }
  $\mathcal{D}_{aug} = \mathcal{D}_{aug} \bigcup \mathcal{D}$\;
\end{algorithm}

\vspace{5pt}\noindent\textbf{Training NLL-enhanced substitutes}. With enriched data, we want to train a substitute model to approximate the NLL property for a given identity-pair $(o,t)$ (the subject and victim pair). For this purpose, we first train our substitute model on the original dataset $\mathcal{D}$ and fine-tune the substitute model on the augmented dataset $\mathcal{D}_{aug}$. And the find-tuned model is expected to be closer to or even surpass a better-trained model around the PoIs, thus, which will elevate the transferability of the adversary example our discovered. 

Naturally, the standard ``softmax'' function was used as the objective function to train our substitute models on the original dataset. On the augmented dataset, we built a triplet loss function according to the NLL: for a tuple $(x_a, x^{(\lambda)}, x_b)$, we set $L_{tri}$ as:
\begin{equation}
\notag
L_{tri} = (C_a(\lambda)-\lambda)^2 + (1-\lambda-C_b(\lambda))^2
\end{equation}
where
\begin{equation}
\notag
C_b(\lambda) = 1 - \frac{R(x^{(\lambda)}) \cdot R(x_b)}{\|R(x^{(\lambda)})\|_2 \|R(x_b)\|_2}
\end{equation}

However, $L_{tri}$ can not be used alone, as it just control the relations among $R(x_a)$, $R(x^{(\lambda)})$ and $R(x_b)$ but dose not give any constraints about the absolute values of $R(x_a)$, $R(x^{(\lambda)})$ and $R(x_b)$ neither the relations among different $R(x^{(\lambda)})$ in different tuples. So using only the triplet loss may break some valuable structures that have been learned from the original dataset.

To solve this problem, we study what the $F(x^{(\lambda)})$ should be when the $R(x^{(\lambda)})$ follows the NLL property. Here, we directly give the conclusion and left the detailed deduction in our Appendix~\ref{app:softmax}. For a tuple $(x_a, x^{(\lambda)}, x_b)$, a well-trained DNN model $F^*$ will produce $F^*(x^{(\lambda)})$ in the following form:
\begin{equation}
\begin{array}{c@{\quad}l}
F^*(x^{(\lambda)})_{(i)} &= 
\begin{cases}
(1+\exp(2\beta \lambda - \beta))^{-1} &, i = a \\
1-F^*(x^{(\lambda)})_{(a)} &, i = b \\
0&, others \\
\end{cases}
\end{array}
\end{equation}

\begin{figure}[ht]
	\centering
  \begin{subfigure}{0.4\textwidth}
		\includegraphics[width=\textwidth]{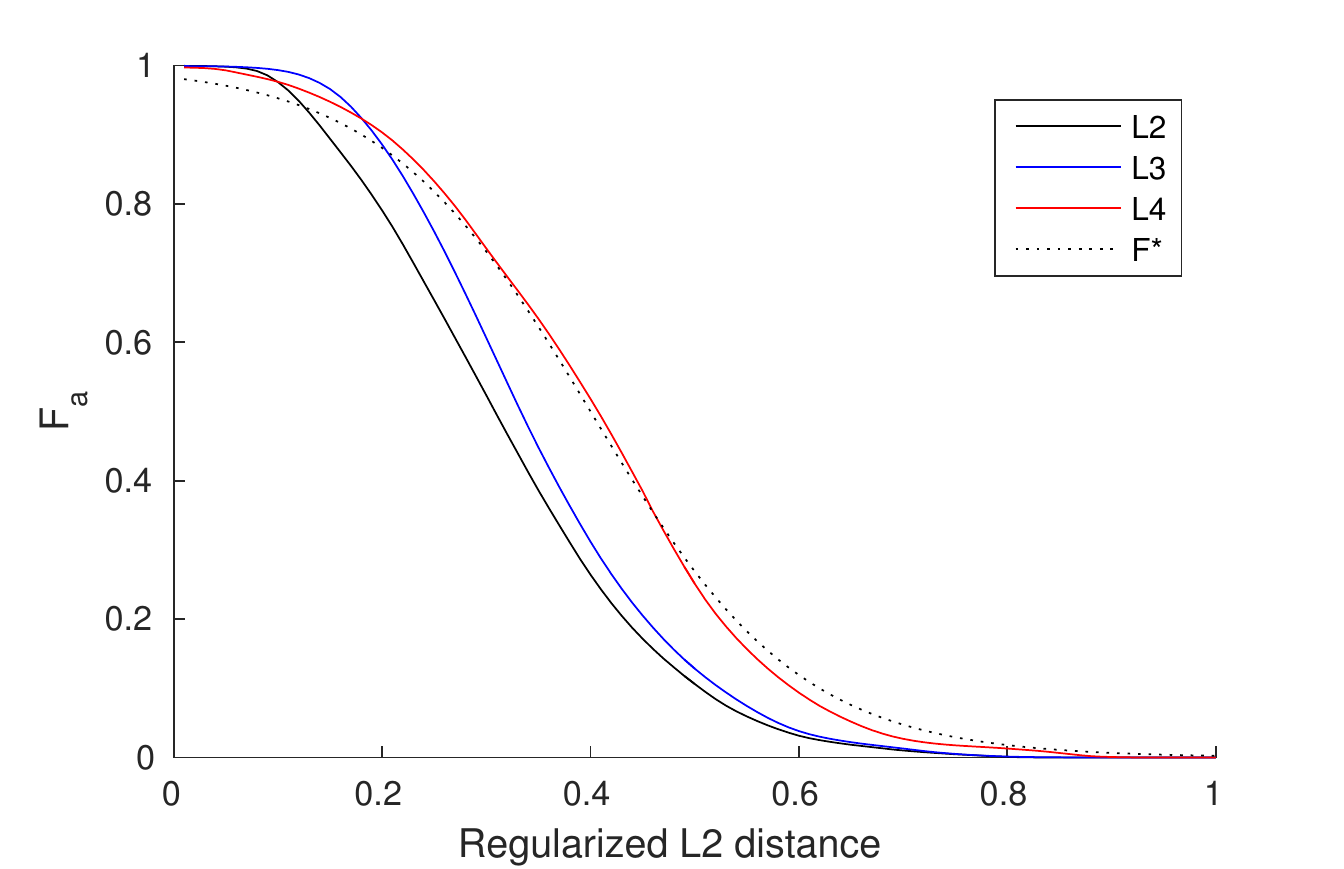}
        
   \end{subfigure}
	\caption{DNN output's patterns: the y-value is the value of the $a$-th element in the output vector of DNN models. }
	\label{fig:rst_soft}

\end{figure}

To determine the value of $\beta$, we tested three ResNet-101 models that trained on L2, L3 and L4 datasets respectively. The results are shown on Fig~\ref{fig:rst_soft}. According to the results, we choose $\beta=4.5$. The curve of $F^*(\cdot)$ when $\beta=4.5$ is also shown on Fig~\ref{fig:rst_soft}.

Combining triplet loss and the results about the expected output of a well-trained DNN, we obtain the objective function for fine-tuning our substitute models as:
\begin{equation}
\begin{array}{c@{\quad}l}
&\sum_{(x_a, x^{(\lambda)}, x_b)} {L_{tri} + L_{soft}(0) + L_{soft}(\lambda) + L_{soft}(1)} \\
\end{array}
\end{equation}
where 
\begin{equation}
\notag
\begin{array}{c@{\quad}l}
& L_{soft}(\lambda) = -\sum_{i}F^*(x^{(\lambda)})_i \log(F(x^{(\lambda)})_i)
\end{array}
\end{equation}
Here, $L_{soft}$ is the Kullback–Leibler divergence between $F(x)$ and $F^*(x)$. And, indeed, $L_{soft}$ can be simplified to contain just two terms about the $a$-th and $b$-th elements of the output vector.

\vspace{5pt}\noindent\textbf{Finding adversarial examples}. After generating multiple substitute models enhanced by the NLL property, we need to effectively assemble them to find transferable adversarial examples. The first question is how to assemble the gradients. The standard approach is to average the gradients discovered from individual substitute model (Eq~\ref{eq:ensemble}). However, we found a way to outperform the standard approach by computing a \textit{weighted} average gradients and clip it according to the agreement. The details are listed in Algorithm 2.

\begin{algorithm}[htb]
\label{alg:gradients}
\caption{Gradients assembling algorithm.}
  \KwIn{$\{R^{(k)}(\cdot)\}$, $w$, $x_t$}
  \KwOut{$\tilde{g}$}
  $x' = tanh(w)$\;
  $K = |\{R^{(k)}(\cdot)\}|$\;
  \For{k = 1 to $K$}{
    $\alpha_k = 1-cos(R^{(k)}(x'),R^{(k)}(x_t))$\;
    $g_k = - \|R^{(k)}(x')\|_2 \bigtriangledown_w {cos(R^{(k)}(x'),R^{(k)}(x_t))}$\;
  }
  $\tilde{g} = \sum_k{\alpha_k g_k} / \sum_k{\alpha_k}$\;
  \For{i = 1 to Dim(w)}{
    $set_{(i)}=\{g_{1(i)}, g_{2(i)}, ..., g_{K(i)}\}$\;
    $p_i = mean(set_{(i)})$\;
    $q_i = std(set_{(i)})$\;
  }
  $max_p = max(\{p_i\})$\;
  \For{i = 1 to Dim(w)}{
    \If{$p_i^2 / q_i \leq 0.3 max_p$}{
      $\tilde{g}_{(i)} = 0$\;
    }
  }
\end{algorithm}

Based on previous knowledge, if an adversarial example which representation is very close, in the terms of Cosine distance, to the representation of the target image, this example will be classified as the target's identity. So, we want to find an adversarial example that can simultaneously shrink the Cosine distance to a vary small value in all of our substitute models. Thus we make heavier for those weights of models in which current modification can hardly shrink the Cosine distance and loose those weights of models in which the current modification has already got a small Cosine distance. This weighting operation have been done in the part before the $7$-th line of the algorithm. The rest of the algorithm tries to clip those gradients in such dimensions where our substitute models do not agree on the direction. More specifically, in these dimensions the variance of gradients from our substitutes is large or the mean is small. In the favor of large mean value, we use $mean(\cdot)^2/std(\cdot)$ to measure the agreement in every dimension and set the bar to be $0.3max_p$.

To be noticed is that the derivate of Cosine distance is:
\begin{equation}
\notag
\bigtriangledown_x{cos(x,x')} = \frac{1}{\|x\|_2} (\frac{x'}{\|x'\|_2}-\frac{x}{\|x\|_2}cos(x,x'))
\end{equation}
So, in our searching algorithm, we regularized the derivate by multiplying the corresponding $L_2$ norm, $\|R^{(k)}(x')\|_2$ (the $5$-th line), which can accelerate the decreasing of the Cosine distance. 

The second question is how to find an adversarial example with small modification. To solve this problem, we designed the Algorithm 3, inspired by the success of multi-step searching algorithm.

\begin{algorithm}[htb]
\label{alg:adv}
\caption{Searching adversarial example algorithm.}
  \KwIn{$K$, $\{R^{(k)}(\cdot)\}$, $x_o$, $x_t$}
  \KwOut{$x'$}
  $\delta = 1$\;
  $w = arctanh(x_o)$\;
  $\bar{c} = K^{-1} \sum_k{cos(R^{(k)}(tanh(w)), R^{(k)}(x_t))}$ \;
  \While{$\bar{c} < 0.8$}{
    \For{i = 1 to $\Theta$} {
      $\Delta = \|tanh(w)-x_o\|_2$ \;
      $g_w = \bigtriangledown_w {\Delta}$ \;
      $g_c = \tilde{g}(w, x_t, \{R^{(k)}(\cdot)\})$ \;
      $g = \exp(\Delta- \delta)g_w + \exp(3-\bar{c})g_c$ \;
      $w = w-lr \times g$\;
      update $\bar{c}$ \;
    }
    $\delta++$ \;
  }
  $x' = tanh(w)$\;
\end{algorithm}

Our searching algorithm is a multi-step algorithm that enlarges the modification limitation $\delta$ step by step. In each step, the algorithm tries $\Theta=1000$ times to find the optimal solution $w$ to minimize the following objective function:  
\begin{equation}
\begin{array}{c@{\quad}l}
 & \exp(\Delta - \delta) + \eta \cdot \exp(1-\bar{c})
\end{array}
\end{equation}
where $\eta$ we chose is $\exp(2)$ which ensure that even in the extreme case where $\bar{c} = 1$ and $\Delta=\delta$ the parameter in front of $g_c$ is still about 7 times larger than the parameter in front of $g_w$. Another advantage of the objective function is that it enforces the algorithm fast decreasing the $1-\bar{c}$ when $\Delta$ is small and bouncing around border when $\Delta$ is near to the modification limitation.

\subsection{Analysis}
\label{subsec:analysis}

To find out how EXCIT enhances the transferability in a query-free, black-box attack, we analyzed our implementation using a ResNet-101 model trained on L3 dataset as the target model, and a set of ResNet-101 models trained on L2 dataset (no overlap with the target's training set except the subjects and victims) as substitute models. Specifically, in the dodging attacks, the attacking set we used is the same with the attacking set mentioned in Section~\ref{subsec:structure}, that contains 100 identities and their 635 images. In the impersonation attacks, we selected 600 image-pairs from 10 identity-pairs (subject-victim). For all identities involved in impersonation attacks, we ensure that each of them have at least 100 images in the Megaface dataset. Under both attack settings, four substitute models were trained using EXCIT method and on augmented dataset, and the target model was attacked by the adversarial example found by these four substitute models. We compare the results with the results of previous studies (Section~\ref{sec:unsderstand}) and the analysis is following.

\begin{figure}[ht]
	\centering
  \begin{subfigure}{0.23\textwidth}
		\includegraphics[width=\textwidth]{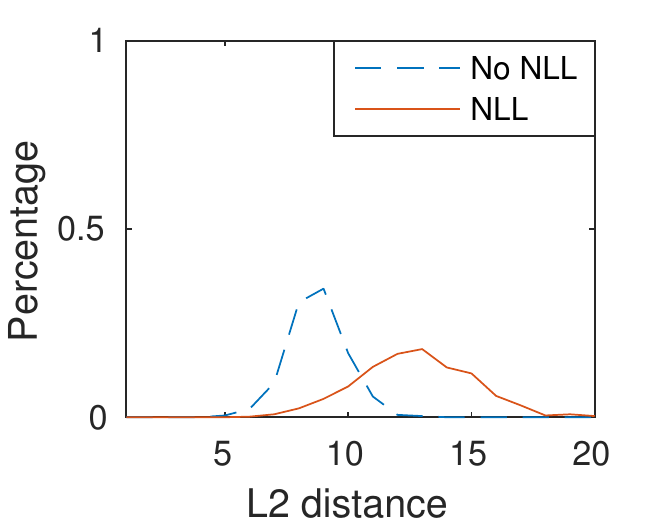}

  \end{subfigure}
  \begin{subfigure}{0.23\textwidth}
		\includegraphics[width=\textwidth]{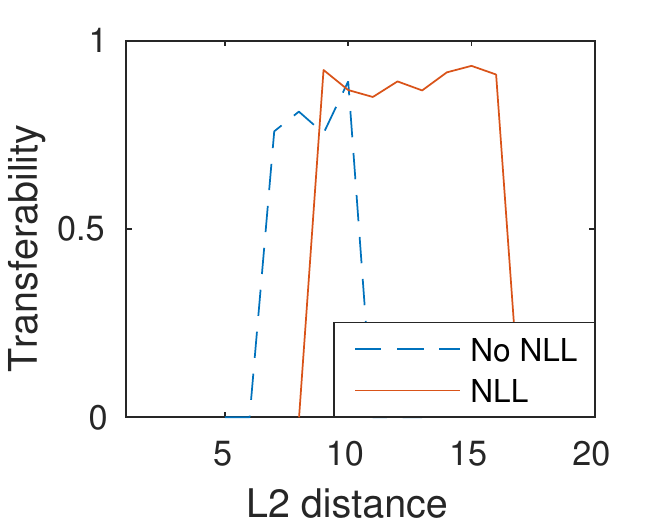}

  \end{subfigure}
	\caption{Dodging performance: The left figure shows the distribution of modifications made by approaches with and without NLL enhance. The right figure shows the transferabilities of them.}
	\label{fig:rst_nll_10w_30w:dod}
\end{figure}

\begin{figure}[ht]
	\centering
  \begin{subfigure}{0.23\textwidth}
		\includegraphics[width=\textwidth]{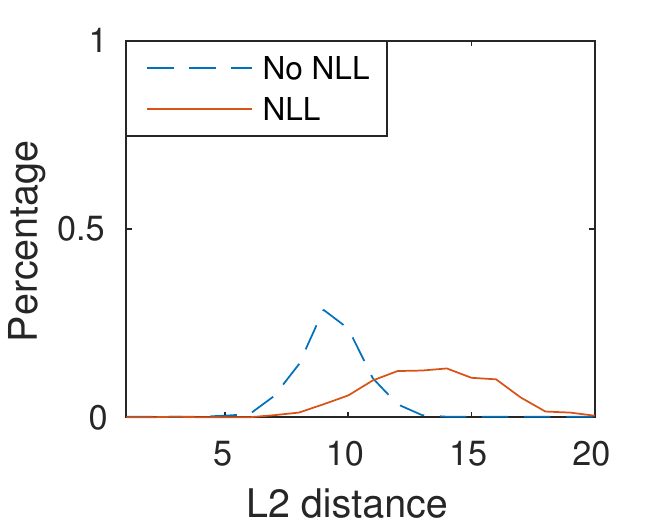}

  \end{subfigure}
  \begin{subfigure}{0.23\textwidth}
		\includegraphics[width=\textwidth]{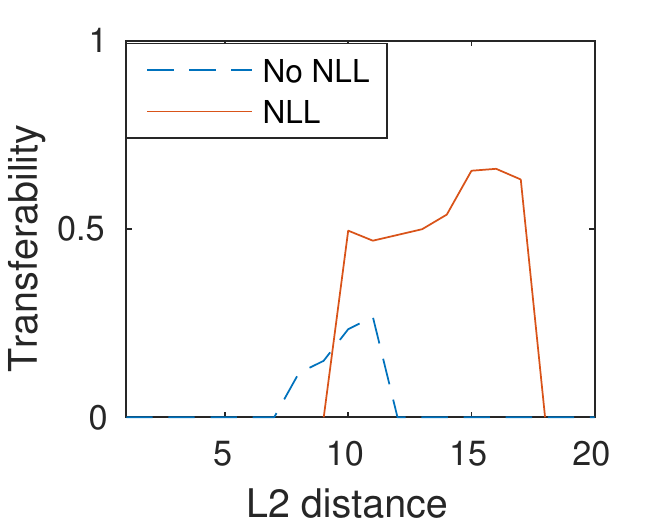}

  \end{subfigure}
	\caption{Impersonation performance: The left figure shows the distribution of modifications made by approaches with and without NLL enhance. The right figure shows the transferabilities of them.}
	\label{fig:rst_nll_10w_30w:imp}
\end{figure}

\vspace{5pt}\noindent\textbf{Transferability}. As we can see from Fig~\ref{fig:rst_nll_10w_30w:dod} and Fig~\ref{fig:rst_nll_10w_30w:imp}, in both attacks, EXCIT improved the transferability, which were evident for dodging (from 81.2\% to 89.6\%) and dramatic for impersonation (from 17.5\% to above 49.8\%). Interesting here is that for the dodging attack, our approach is close to the attack using the substitute models trained on the same level dataset with the target model, indicating that our EXCIT model is actually effective in recognizing the subject. This is further supported by the findings for the impersonation attack, in which none of the substitute models without the NLL enhancement could come even close to our performance, even for those as well-trained as the target model. Actually, even for a ResNet-101 trained on L3 dataset, we found that our substitute models got a transferability of 49.8\%. 

\begin{table}[tbh]
  \centering
    \caption{Impersonation transferability of NLL-enhanced approach on different structures. The number in the bracket is the transferability using substitute models trained on L3 datasets.}
\scalebox{0.9}[0.9]{
	\begin{tabular}{|c|c|c|c|}
	\hline
	  & ResNet-101 & GoogLeNet & VGG-16\\
    \hline
	with NLL & 49.8\% (72.1\%) &  46.8\% & 43.5\% \\
	\hline
    without NLL & 17.5\% (25.2\%) &  16.2\% & 16.3\% \\
	\hline
	\end{tabular}
	}
	\label{tb:nll}

\end{table}

Further our study shows that EXCIT also works on other DNN structures: we use the same setting (four L2 substitute models to attack L3 target model) to test the effectivity of EXCIT on VGG-16 and GoogLeNet structures, and found that (Table~\ref{tb:nll}) both attacks achieved around 45\% transferability, way above the 16\% reported in previous data-size study (Section~\ref{subsec:size}).

\vspace{5pt}\noindent\textbf{Distance from the subject}. Without querying the target, naturally the adversarial examples discovered by EXCIT tend to be farther away from the original image. What we want to know, however, is for a given distance from the subject, whether the adversarial examples found by our approach still have a higher probability of success, compared with the attack without the NLL enhancement. For this purpose, we use four substitute models trained on L2 datasets to attack the target model trained on L3 dataset, and compare both methods' average transferability under various $L_2$ distances. The results are given in Table~\ref{tb:t_dist}. A more detailed study, by restricting the searching space within a certain radius is given in the Appendix~\ref{app:dist}. From these results, we observe that NLL improves the transferability in every distance. 
\begin{table}[tbh]

  \centering
    \caption{Transferability under different distance constraints.}
\scalebox{0.9}[0.9]{
	\begin{tabular}{|c|c|c|c|}
	\hline
	  & $<10$ & $<15$ & $<20$\\
    \hline
	Dodging with NLL & 87.7\% &  88.3\% & 89.6\% \\
	\hline
    Dodging without NLL& 79.5\% &  81.2\% & 81.2\% \\
	\hline
	Impersonation with NLL& 50.5\% &  47.2\% & 49\% \\
	\hline
    Impersonation without NLL& 14.5\% & 17.5\% & 17.5\% \\
	\hline
	\end{tabular}
    }
	\label{tb:t_dist}

\end{table}

\vspace{5pt}\noindent\textbf{Training cost}. To understand how EXCIT helps a resource-limited adversary, we evaluated the training cost of substitute models over our 8-GPU server (with the 12GB memory for each GPU). As illustrated in Table~\ref{tb:cost}, it took about 200 hours to train a ResNet-101 on L3 dataset and more than 500 hours to build the model on L4 dataset, while constructing a substitute models on L2 dataset used 75 hours. Note that we could fully parallelize the training of 4 substitute models, but could not do this for a target model over the same amount of computing resources, due to the communication overheads.

To further analyze the cost EXCIT could reduce, we trained four substitute models of ResNet-50 and four of ResNet-101 on L1 datasets to perform impersonation attacks against the target ResNet-101 models trained on L3 and L4 datasets. As we can see in Tabel~\ref{tb:r50}, with the small amount of training data, our approach elevated the transferability of these attacks to 30-40\% for target model trained on L3 datasets and around 20\% for the target model trained on L4 dataset, while the training time stayed at 5 to 9 hours per substitute model. 

\begin{table}[tbh]

  \centering
    \caption{Impersonation transferability with NLL-enhanced approach using L1 models as substitutes. Cell $(i,j)$ is the result of using model $i$ to attack model $j$.}
	\begin{tabular}{|c|c|c|}
	\hline
	  & L3 & L4\\
    \hline
	ResNet-50 L1 & 38.5\% & 16.8\% \\
	\hline
    ResNet-101 L1 & 44.2\% & 23.2\% \\
	\hline
	\end{tabular}
	
	\label{tb:r50}

\end{table}

Also we compared the efficiency of our approach against the attacks without the NLL enhancement. In the latter case, the only way to improve the transferability is to train more substitute models to find their common adversary examples. In our study, we built three experiments using 4, 8 and 16 substitute models respectively, each model trained on a L1 dataset with ResNet-101 structure. In these experiments, the target model is trained on L4 dataset and also with ResNet-101 structure. The results of the standard ensemble method over these models are presented in Table~\ref{tb:en_models}. As we can see here, when attacking the L4 target, even with 16 substitutes, the attack could not achieve the same level of transferability as the 4 NLL-enhanced substitutes, even the substitutes with only 50 layers. In this case, the cost of EXCIT, in terms of training time, is no more than 16.8\% of the direct attack (with 16 substitutes).

\begin{table}[tbh]

  \centering
    \caption{Impersonation transferability of standard ensemble method using four substitute models (no NLL) trained on L1 datasets.}
	\begin{tabular}{|c|c|c|}
	\hline
	  4 models & 8 models & 16 models\\
    \hline
	7.2\% & 12.7\% & 16.5\% \\
	\hline
	\end{tabular}
	
	\label{tb:en_models}

\end{table}

\begin{table}[tbh]
  \centering
    \caption{Cost of training different models.}
	\begin{tabular}{|c|c|c|c|}
	\hline
	 Depth & Dataset & Time &  Memory\\
    \hline
	50 & L1 & 5h & 8x4.5G \\
	\hline
    101 & L1 & 9h & 8x6.5G \\
	\hline
    101 & L2 & 75h & 8x7.5G \\
	\hline
    101 & L4 & >500h & 8x8G\\
	\hline
	\end{tabular}
	
	\label{tb:cost}
\end{table}

\section{Evaluation on Real-World Systems}
\label{sec:evaluation}

We evaluated our approach, EXCIT, on four real world systems. Three of them are online, with APIs available for the public, and the last one is a commercial system without open access, one of the products from SenseTime Ltd. We performed both dodging and impersonation attacks against them. The details of our experiments and our findings are elaborated below.

\subsection{Experimental Settings}
\label{subsec:setting}

Unlike the models built in our analysis, which were trained over the subject and victim's images and output a vector specifying the possibility of the input image belongs to every identity, a real world FR system takes two photos as inputs and calculates a score about the similarity of the individuals in these two photos. Here is how we determined whether an adversarial example worked on the real world FR systems: 

In a successful dodging attack, we expect that the target system outputs a low score ($< Th_{dod}$) for two images: one is the subject's original photo and the other is the adversarial example generated by our approach from the original photo. In a successful impersonation attack, the target system is supposed to output a high score ($> Th_{imp}$) for two images: the victim's photo and the adversarial example generated by our approach from the subject's image, indicating that they are belong to the same individual. As usual, the thresholds $Th_{dod}$ and $Th_{imp}$ are specified by the FR system.

In our experiments, we first trained 4 ResNet-101 models on randomly sampled 3M photos from 600K identities in the MegaFace Challenge 2 dataset\footnote{We did not use all 4M for each substitute in an attempt to make these substitutes diverse.}.

From all identities, we selected 10 individuals as the subjects in our dodging attack. For the impersonation, we sampled 10 subject-victim pairs. Every identity involved has at least 100 images in the Megaface dataset. In the experiments, we randomly chose 10 of each individual's images for the dodging attack and 10 photo-pairs for each subject-victim pair to execute the impersonation attack. For each of these subjects or subject-victim pairs, we used our method to augment the dataset and fine-tuned the substitute models on this augmented dataset.

\subsection{Attack on Online APIs}
\label{subsec:api}

The three online APIs attacked in our research are ColorReco\footnote{http://www.colorreco.com/faceCompare}, FaceVisa\footnote{http://www.facevisa.com/web/index/demo} and Face++\footnote{https://www.faceplusplus.com/face-comparing/\#demo}.  The models behind these APIs were trained with a large amount of data. For example, Face++ was built upon 5M photos of 20K identities~\cite{zhou2015naive} and FaceVisa was upon 2M photos. Also they all demonstrated a high recognition accuracy over the Labeled Faces in the Wild (LFW) dataset~\cite{huang2007labeled}: $99.4\%$ for ColorReco, $99.5\%$ for FaceVisa and $99.5\%$ for Face++. In our experiments, we ran a python script to automatically upload our test photo pairs to ColorReco and FaceVisa. For Face++, we had to do it manually due to the requirement of CAPTCHA solving. 

\begin{figure}[ht]
	\centering
  \begin{subfigure}{0.15\textwidth}
		\includegraphics[width=\textwidth]{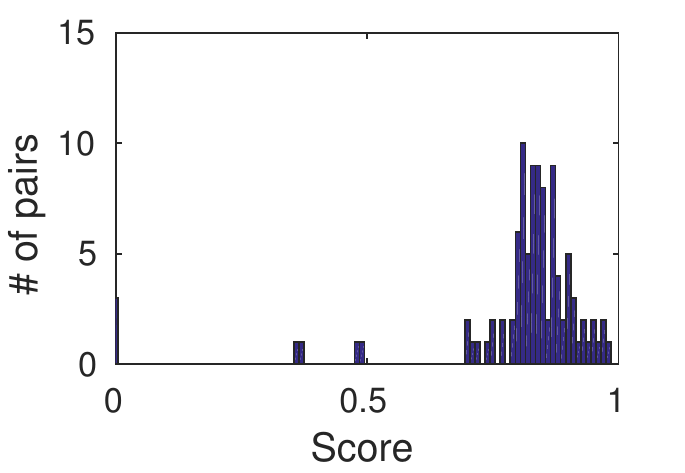}

  \end{subfigure}
  \begin{subfigure}{0.15\textwidth}
		\includegraphics[width=\textwidth]{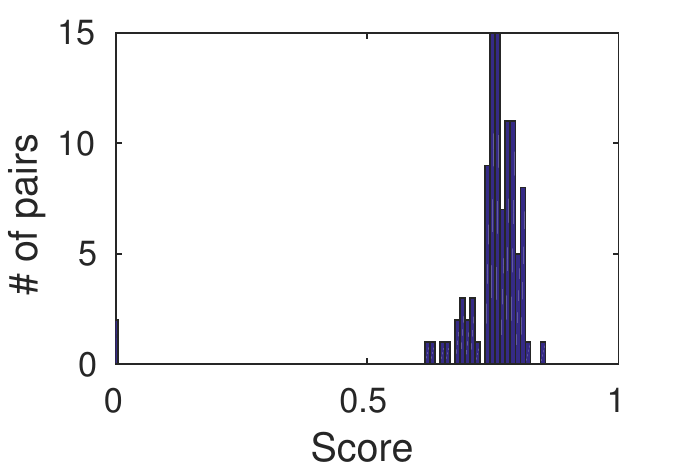}

  \end{subfigure}
  \begin{subfigure}{0.15\textwidth}
		\includegraphics[width=\textwidth]{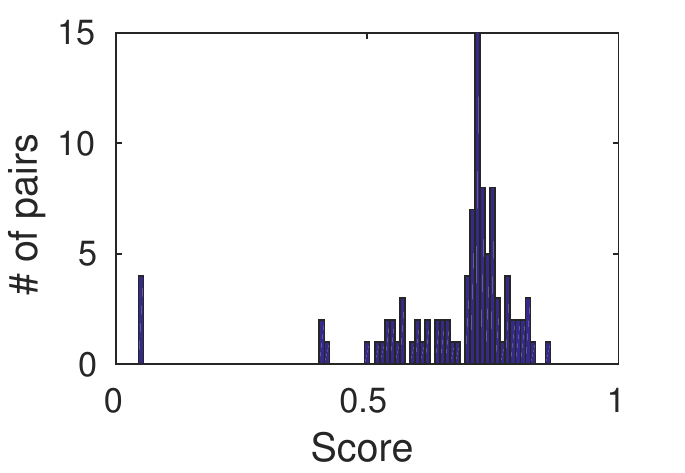}

  \end{subfigure}
 
	\caption{Distributions of scores for impersonation attacks: from left to right, they are the results of ColorReco, Facevisa and Face++ respectively.}
	\label{fig:evl_apis}

\end{figure}

The success rates of our attacks are presented in Table~\ref{tb:apis}. Note that in these experiments, the thresholds for different APIs are different and defined by the APIs themselves. Besides, we set that an attack failed if the target FR system could not detect face from the adversarial example submitted, even for the dodging attack. As we can see from the table, our approach achieved a higher accuracy in the dodging attack, compared with the attacks without the NLL enhancement (Table~\ref{tb:nll}). A much bigger boost, however, is observed for the impersonation attack, in which EXCIT raised the success rates for all three systems from around 20\% to 69-85\%.  Fig~\ref{fig:evl_apis} further illustrates the distributions for the scores of our submitted photo pairs. 

\begin{table}[tbh]
  \centering
    \caption{Success rate against online APIs.}
	\begin{tabular}{|c|c|c|c|}
	\hline
	  & ColorReco & Facevisa &  Face++\\
    \hline
	dodging & 98\% & 96\% & 95\% \\
	\hline
	$Th_{dod}$ & 0.75 & 0.64 & 0.623 \\
	\hline
	impersonation & 74\% & 85\% & 69\% \\
	\hline
	$Th_{imp}$ & 0.80 & 0.74 & 0.691 \\
	\hline
	
	\end{tabular}
	
	\label{tb:apis}

\end{table}

\subsection{Attack on Industrial System}
\label{subsec:realsystem}

\begin{figure}[ht]
	\centering
  \begin{subfigure}{0.15\textwidth}
		\includegraphics[width=\textwidth]{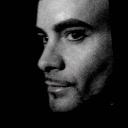}
  \end{subfigure}
  \begin{subfigure}{0.15\textwidth}
		\includegraphics[width=\textwidth]{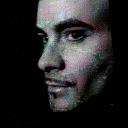}
  \end{subfigure}
  \begin{subfigure}{0.15\textwidth}
		\includegraphics[width=\textwidth]{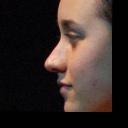}
  \end{subfigure}
  
  \begin{subfigure}{0.15\textwidth}
		\includegraphics[width=\textwidth]{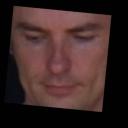}
  \end{subfigure}
  \begin{subfigure}{0.15\textwidth}
		\includegraphics[width=\textwidth]{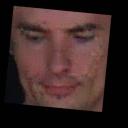}
  \end{subfigure}
  \begin{subfigure}{0.15\textwidth}
		\includegraphics[width=\textwidth]{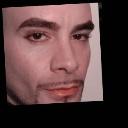}
  \end{subfigure}
 
	\caption{Successful impersonation attacks on SenseTime. The three columns are the subject photos, our generated adversarial examples and the target photos respectively. The modification of the first case is 13.41 and the second case is 7.22.}
	\label{fig:evl_st}

\end{figure}

Commercial FR systems are often better trained and more capable than the free FR APIs, which are mostly used for online demo. Such industry-grade systems are typically characterized by a large number of layers, and being trained over a massive amount of data on clusters of GPUs. The services they provide are not open to the public and only available for purchase. In our research, we obtained the commercial SDKs from SenseTime Ltd. through our collaborations. SenseTime's products are known to be among the leading FR systems~\cite{sun2015deepid3}. So the system we analyzed represents the state-of-the-art in FR technologies. It was trained over 20M photos for 1M individuals, using a ResNet-like model, though the details of the structure are commercial secrets. The model tested in our study was estimated to require at least 14,000 hours (50 epochs) to train, over our GPU server. By comparison, all 4 EXCIT substitutes used in our attack were trained for 2,500 hours in total. With less than 1/5 of the time spent on training the models, our approach achieved a high success rate for the 100 individual selected for the dodge attack and 100 pairs for the impersonation attack: in the former case, 89\% of transferability was achieved, compared to 70\% without the NLL enhancement, and the in the latter, we raised the transferability from 11\% to 62\%. Fig~\ref{fig:evl_st} further shows examples for the successful attacks. We have reported our findings to SenseTime and are helping them improve their system. 

\section{Related Works}
\label{sec:relatedWork}

Our approach utilizes a synthesized dataset to fine-tune our substitutes, for the purpose of approximating the NLL property at PoIs. Synthesized data have also been used in the prior research~\cite{papernot2017practical}, to support completely different techniques and to different purpose. Specifically, the prior research uses synthesized data to query the target model to train the substitute (1000 times to achieve an 84.24\% success rate in transferring adversarial examples).  This is \textit{exact} the attack scenario our \textit{query-free} approach is designed to avoid. Without communicating with the target model, the only thing we can do is to build a substitute as well-trained as the target, so as to captures their common structural weaknesses to enhance transferrability. Such an attack is found to be completely feasible through simulating the target's behaviors around PoIs, based upon the NLL property we discovered and additional images collected from the victims and the attackers.

Our approach also exploited the ensemble method. The original ensemble-based approach is proposed by Liu et al.~\cite{liu2016delving}. Their work focus on the transferability among DNN models with different structures, whereas the data size is the factor that really matters in face recognition domain, as have been demonstrated before. Compared with them, our method can increase the transferability from a model trained with insufficient data to a model trained with plenty of data, which goes beyond their method's capability. Another ensemble method is proposed by Sarkar et al.~\cite{sarkar2017upset}. They trained a DL model to find ``universal'' perturbations fooling all their pre-trained target models. Particularly, their object function combines both the sum of targets' (mis)classification loss and the scale of the finding perturbations. The difference from their method is similar with above. In our attacking scenario, their method may not work.

Concurrent to our work, \textit{mixup}~\cite{zhang2017mixup} also utilizes the interpolated images to augment the dataset for optimizing KL-divergence loss function. However, \textit{mixup} labels $F^*(x^{(\lambda)})$ with $\lambda$, as we can see from Fig~\ref{fig:rst_soft}, which is just a rough approximation of the DNN's real output. In contrast, based on the NLL, we deduce the real formation of the output (see Eq 4) and further improve our data augmentation method using this formation. Also, \textit{mixup} is designed to improve the classification accuracy while our method is meant to transfer the NLL property across models. So we believe that our approach is likely to perform better when it comes to transferability, given our better prediction of the DNN outputs (which \textit{mixup} just uses a linear function to estimate).  

%Even though we have not built experiments to test which method is better to improve the overall performance, we believe, at least in the FR field, our method is better, as we are more close to the real. 

\section{Discussion}
\label{sec:discussion}

\noindent\textbf{Understanding NLL}. Unlike existing cross-model attacks, EXCIT does not even interact with the target, so there is no way for our approach to exploit the specific defects of the target. The reason we can still find highly transferable examples is that by simulating better-trained models around PoIs, our approach is likely to discover some common (potentially structural) defects fundamental to a certainly type of DNN, and in the meantime, avoid exploring the subspace unlikely to contain adversarial examples, given the reduction of training-specific weaknesses (e.g., lack of sufficient data) around PoIs. Under an NLL-enhanced substitute, we could even discover transferable examples with modifications are restricted to a given facial region: e.g., around the eyes (Appendix~\ref{app:eye}), which allows the prior attack~\cite{sharif2016accessorize} (using printout glasses to evade detection) to work in a query-free, black-box setting. 

In the meantime, our understanding of transferability is still limited. Still less clear are the questions such as whether there exist adversarial examples inherent to certain DNN structures or even the fundamental design of artificial neural networks. Further studies on these issues are certainly important. 

Our current definition of NLL describes a relation between Cosine distance and the $L_2$ distance. However, such a relation may not be general, particularly when it comes to non-FR problems: as an example, we found that although the NLL-enhanced models still improve transferability over Cifar10~\cite{krizhevsky2009learning}, a dataset for image classification, the enhancement is less significant (Table~\ref{tb:cifar10}). This could be attributed to the unique features of FR: e.g., differences between two faces can be added to another face to form a new face, which makes the ``transitional'' images easy to construct; also FR tasks are characterized by a large number of training categories, compared with other tasks (e.g., 1K categories for ILSVRC vs. 600K entities for Megaface), which forces the DNN model to map input images to a high-dimensional sphere with a maximum space utilization. All these features make NLL more effective on the FR tasks. What is less clear, however, is how to extend the concept to improve the transferability of other tasks, which should be studied in future research. 

\begin{table}[tbh]
  \centering
    \caption{Transferability on Cifar10. We used 4 substitutes (ResNet-20) trained on 10K images to attack the target model (ResNet-20) trained on 30K images.}
	\begin{tabular}{|c|c|c|}
	\hline
	  & Dodging & Impersonation \\
    \hline
	with & 100\% & 57\%\\
	\hline
	without & 100\% & 63\% \\
	\hline
	\end{tabular}
	
	\label{tb:cifar10}

\end{table}

%\todo{Please double-check. Make sure that I said the right things. }

\vspace{5pt}\noindent\textbf{Defense}. Defensive $distillation$~\cite{papernot2016distillation} has been demonstrated to be effective against most of previous attacks. However, as pointed out by Carlini et al~\cite{carlini2017towards}, a modified version of existing attacks will break them defense. We also found that $distillation$ does not work on EXCIT either: more specifically, We implemented the defense on our L3 target model with temperature $T = 20$, and ran four L2 substitute models to attack it. The result is that $distillation$ can only reduce our transferability from $89.6\%$ to $88.8\%$, for dodging, and from $49\%$ to $45.6\%$, for impersonation.

Alternatively, we can consider to insert ``secret'' into the commercial system. Specifically, train the system on a custom dataset where all the photos are covered by a secret pattern. Since queries are not supposed to be made to the target during an attack, the secret added to a commercial system could help mitigate the EXCIT threat. In general, however, defense against adversarial learning is known to be hard~\cite{carlini2017adversarial}. Further research is needed to find an effective way to defeat our attack.

\vspace{5pt}\noindent\textbf{Cost of the attack}. As mentioned earlier, training the substitutes to attack SenseTime's FR system took 2,500 hours on our server. An estimate cost for such resources is about 10000 dollars on Amazon AWS. The computing time here can be shortened through parallelization, since all 4 substitutes can be trained together. Also, the computing cost could be reduced when the adversary attempts to impersonate multiple victims or hide multiple subjects. In this case, only one set of substitutes need to be trained over our dataset, which can later be augmented with NLL for different subjects or subject-victim pairs to support dodging or impersonation attacks.

\section{Conclusion}
\label{sec:conclusion}
In this paper, we present our new understanding of DNN-based FR systems, in terms of their vulnerability to the transferable attack under a resource-constrained adversary. Our research shows that limited resources, particularly smaller training sets, can have a significant impact on the effectiveness of the attack. This is important since a real-world adversary typically cannot query the target frequently and  needs to build substitutes as capable as, or even more powerful than the target model under his limited resources. Narrowing such a resource gap, however, turns out to be feasible through a novel technique we developed. Specifically, we found that the adversary could make an effective use of the extra information (images) about subjects and victims in his possession, by approximating the relations of these PoIs with other images in the training set characterized by a Near Local Linearity (NLL) property we discovered.  As a result, we can grossly elevate the transferability in both a dodging and an impersonation attack by training NLL-enhanced models by nearly 50\% in attacking industry-grade systems. With our new techniques and findings, still more effort needs to be made to better understand transferability and mitigate the threat it poses.

\bibliographystyle{IEEEtranS}
\bibliography{main}

% Generated by IEEEtranS.bst, version: 1.12 (2007/01/11)
\begin{thebibliography}{10}
\providecommand{\url}[1]{#1}
\csname url@samestyle\endcsname
\providecommand{\newblock}{\relax}
\providecommand{\bibinfo}[2]{#2}
\providecommand{\BIBentrySTDinterwordspacing}{\spaceskip=0pt\relax}
\providecommand{\BIBentryALTinterwordstretchfactor}{4}
\providecommand{\BIBentryALTinterwordspacing}{\spaceskip=\fontdimen2\font plus
\BIBentryALTinterwordstretchfactor\fontdimen3\font minus
  \fontdimen4\font\relax}
\providecommand{\BIBforeignlanguage}[2]{{%
\expandafter\ifx\csname l@#1\endcsname\relax
\typeout{** WARNING: IEEEtranS.bst: No hyphenation pattern has been}%
\typeout{** loaded for the language `#1'. Using the pattern for}%
\typeout{** the default language instead.}%
\else
\language=\csname l@#1\endcsname
\fi
#2}}
\providecommand{\BIBdecl}{\relax}
\BIBdecl

\bibitem{carlini2017adversarial}
N.~Carlini and D.~Wagner, ``Adversarial examples are not easily detected:
  Bypassing ten detection methods,'' \emph{arXiv preprint arXiv:1705.07263},
  2017.

\bibitem{carlini2017towards}
------, ``Towards evaluating the robustness of neural networks,'' in
  \emph{Security and Privacy (SP), 2017 IEEE Symposium on}.\hskip 1em plus
  0.5em minus 0.4em\relax IEEE, 2017, pp. 39--57.

\bibitem{fan2014learning}
H.~Fan, Z.~Cao, Y.~Jiang, Q.~Yin, and C.~Doudou, ``Learning deep face
  representation,'' \emph{arXiv preprint arXiv:1403.2802}, 2014.

\bibitem{goodfellow2015explaining}
I.~J. Goodfellow, J.~Shlens, and C.~Szegedy, ``Explaining and harnessing
  adversarial examples,'' \emph{stat}, vol. 1050, p.~20, 2015.

\bibitem{he2016deep}
K.~He, X.~Zhang, S.~Ren, and J.~Sun, ``Deep residual learning for image
  recognition,'' in \emph{Proceedings of the IEEE conference on computer vision
  and pattern recognition}, 2016, pp. 770--778.

\bibitem{huang2007labeled}
G.~B. Huang, M.~Ramesh, T.~Berg, and E.~Learned-Miller, ``Labeled faces in the
  wild: A database for studying face recognition in unconstrained
  environments,'' Technical Report 07-49, University of Massachusetts, Amherst,
  Tech. Rep., 2007.

\bibitem{huang2017adversarial}
S.~Huang, N.~Papernot, I.~Goodfellow, Y.~Duan, and P.~Abbeel, ``Adversarial
  attacks on neural network policies,'' \emph{arXiv preprint arXiv:1702.02284},
  2017.

\bibitem{ioffe2015batch}
S.~Ioffe and C.~Szegedy, ``Batch normalization: Accelerating deep network
  training by reducing internal covariate shift,'' in \emph{International
  Conference on Machine Learning}, 2015, pp. 448--456.

\bibitem{jia2014caffe}
Y.~Jia, E.~Shelhamer, J.~Donahue, S.~Karayev, J.~Long, R.~Girshick,
  S.~Guadarrama, and T.~Darrell, ``Caffe: Convolutional architecture for fast
  feature embedding,'' in \emph{Proceedings of the 22nd ACM international
  conference on Multimedia}.\hskip 1em plus 0.5em minus 0.4em\relax ACM, 2014,
  pp. 675--678.

\bibitem{krizhevsky2009learning}
A.~Krizhevsky and G.~Hinton, ``Learning multiple layers of features from tiny
  images,'' 2009.

\bibitem{krizhevsky2012imagenet}
A.~Krizhevsky, I.~Sutskever, and G.~E. Hinton, ``Imagenet classification with
  deep convolutional neural networks,'' in \emph{Advances in neural information
  processing systems}, 2012, pp. 1097--1105.

\bibitem{lecun1998gradient}
Y.~LeCun, L.~Bottou, Y.~Bengio, and P.~Haffner, ``Gradient-based learning
  applied to document recognition,'' \emph{Proceedings of the IEEE}, vol.~86,
  no.~11, pp. 2278--2324, 1998.

\bibitem{liu2016delving}
Y.~Liu, X.~Chen, C.~Liu, and D.~Song, ``Delving into transferable adversarial
  examples and black-box attacks,'' \emph{arXiv preprint arXiv:1611.02770},
  2016.

\bibitem{nech2017level}
A.~Nech and I.~Kemelmacher-Shlizerman, ``Level playing field for million scale
  face recognition,'' in \emph{Proceedings of the IEEE Conference on Computer
  Vision and Pattern Recognition}, 2017.

\bibitem{papernot2016effectiveness}
N.~Papernot and P.~McDaniel, ``On the effectiveness of defensive
  distillation,'' \emph{arXiv preprint arXiv:1607.05113}, 2016.

\bibitem{papernot2016transferability}
N.~Papernot, P.~McDaniel, and I.~Goodfellow, ``Transferability in machine
  learning: from phenomena to black-box attacks using adversarial samples,''
  \emph{arXiv preprint arXiv:1605.07277}, 2016.

\bibitem{papernot2017practical}
N.~Papernot, P.~McDaniel, I.~Goodfellow, S.~Jha, Z.~B. Celik, and A.~Swami,
  ``Practical black-box attacks against machine learning,'' in
  \emph{Proceedings of the 2017 ACM on Asia Conference on Computer and
  Communications Security}.\hskip 1em plus 0.5em minus 0.4em\relax ACM, 2017,
  pp. 506--519.

\bibitem{papernot2016limitations}
N.~Papernot, P.~McDaniel, S.~Jha, M.~Fredrikson, Z.~B. Celik, and A.~Swami,
  ``The limitations of deep learning in adversarial settings,'' in
  \emph{Security and Privacy (EuroS\&P), 2016 IEEE European Symposium
  on}.\hskip 1em plus 0.5em minus 0.4em\relax IEEE, 2016, pp. 372--387.

\bibitem{papernot2016distillation}
N.~Papernot, P.~McDaniel, X.~Wu, S.~Jha, and A.~Swami, ``Distillation as a
  defense to adversarial perturbations against deep neural networks,'' in
  \emph{Security and Privacy (SP), 2016 IEEE Symposium on}.\hskip 1em plus
  0.5em minus 0.4em\relax IEEE, 2016, pp. 582--597.

\bibitem{ILSVRC15}
O.~Russakovsky, J.~Deng, H.~Su, J.~Krause, S.~Satheesh, S.~Ma, Z.~Huang,
  A.~Karpathy, A.~Khosla, M.~Bernstein, A.~C. Berg, and L.~Fei-Fei, ``{ImageNet
  Large Scale Visual Recognition Challenge},'' \emph{International Journal of
  Computer Vision (IJCV)}, vol. 115, no.~3, pp. 211--252, 2015.

\bibitem{sarkar2017upset}
S.~Sarkar, A.~Bansal, U.~Mahbub, and R.~Chellappa, ``Upset and angri: Breaking
  high performance image classifiers,'' \emph{arXiv preprint arXiv:1707.01159},
  2017.

\bibitem{schroff2015facenet}
F.~Schroff, D.~Kalenichenko, and J.~Philbin, ``Facenet: A unified embedding for
  face recognition and clustering,'' in \emph{Proceedings of the IEEE
  Conference on Computer Vision and Pattern Recognition}, 2015, pp. 815--823.

\bibitem{sharif2016accessorize}
M.~Sharif, S.~Bhagavatula, L.~Bauer, and M.~K. Reiter, ``Accessorize to a
  crime: Real and stealthy attacks on state-of-the-art face recognition,'' in
  \emph{Proceedings of the 2016 ACM SIGSAC Conference on Computer and
  Communications Security}.\hskip 1em plus 0.5em minus 0.4em\relax ACM, 2016,
  pp. 1528--1540.

\bibitem{simonyan2014very}
K.~Simonyan and A.~Zisserman, ``Very deep convolutional networks for
  large-scale image recognition,'' \emph{arXiv preprint arXiv:1409.1556}, 2014.

\bibitem{sun2015deepid3}
Y.~Sun, D.~Liang, X.~Wang, and X.~Tang, ``Deepid3: Face recognition with very
  deep neural networks,'' \emph{arXiv preprint arXiv:1502.00873}, 2015.

\bibitem{szegedy2015going}
C.~Szegedy, W.~Liu, Y.~Jia, P.~Sermanet, S.~Reed, D.~Anguelov, D.~Erhan,
  V.~Vanhoucke, and A.~Rabinovich, ``Going deeper with convolutions,'' in
  \emph{Proceedings of the IEEE conference on computer vision and pattern
  recognition}, 2015, pp. 1--9.

\bibitem{szegedy2013intriguing}
C.~Szegedy, W.~Zaremba, I.~Sutskever, J.~Bruna, D.~Erhan, I.~Goodfellow, and
  R.~Fergus, ``Intriguing properties of neural networks,'' \emph{arXiv preprint
  arXiv:1312.6199}, 2013.

\bibitem{taigman2014deepface}
Y.~Taigman, M.~Yang, M.~Ranzato, and L.~Wolf, ``Deepface: Closing the gap to
  human-level performance in face verification,'' in \emph{Proceedings of the
  IEEE conference on computer vision and pattern recognition}, 2014, pp.
  1701--1708.

\bibitem{xu2017feature}
W.~Xu, D.~Evans, and Y.~Qi, ``Feature squeezing: Detecting adversarial examples
  in deep neural networks,'' \emph{arXiv preprint arXiv:1704.01155}, 2017.

\bibitem{zhang2017mixup}
H.~Zhang, M.~Cisse, Y.~N. Dauphin, and D.~Lopez-Paz, ``mixup: Beyond empirical
  risk minimization,'' \emph{arXiv preprint arXiv:1710.09412}, 2017.

\bibitem{zhou2015naive}
E.~Zhou, Z.~Cao, and Q.~Yin, ``Naive-deep face recognition: Touching the limit
  of lfw benchmark or not?'' \emph{arXiv preprint arXiv:1501.04690}, 2015.

\end{thebibliography}

\begin{appendices}

\section{Transferability prediction}
\label{app:transferability}

Given an adversarial example discovered, the attacker needs to have some idea how likely the example could also mislead the target model. Also in the presence of multiple examples, the most promising one would be given preference. One way to estimate the transferability of an example is to train multiple models as capable as the target, called \textit{target simulators} or simply \textit{simulators}, run substitutes to attack them and then collect the statistics about the relation between the features of the adversarial examples discovered in substitutes and the transferability of the examples. Given such a relation, the attacker can look at the features of an example to estimate the likelihood that it could fool the target. This simple approach, however, does not work in practice, as we do \textit{not} have the resources (e.g., a large number of images) to build such powerful simulators.  

Therefore in our research, we took a different path including three steps: firstly, we train simulators on the data we have; secondly, we estimate the difference between simulators and the target, thirdly, we plus the estimated difference to our simulators' outputs to predict the target's output. In these process, estimation the difference is challenge, as we don't know how the target model looks like. But we can know the scale of the training set of the target model. Thus we built a function $g(m_{\alpha},m_{\beta})$ to estimate the difference between models trained on $m_{\alpha}$ images and $m_{\beta}$ images. Specifically, our study shows that when training data grows, the substitute becomes similar to the target, and the impact of their data size difference becomes less prominent. Next, plussing the average difference to every simulator's output, we derive what the target model would output for adversarial examples and can choose the best one with the largest likelihood that it will fool the target. 

\begin{figure}[ht]
	\centering
  \begin{subfigure}{0.23\textwidth}
		\includegraphics[width=\textwidth]{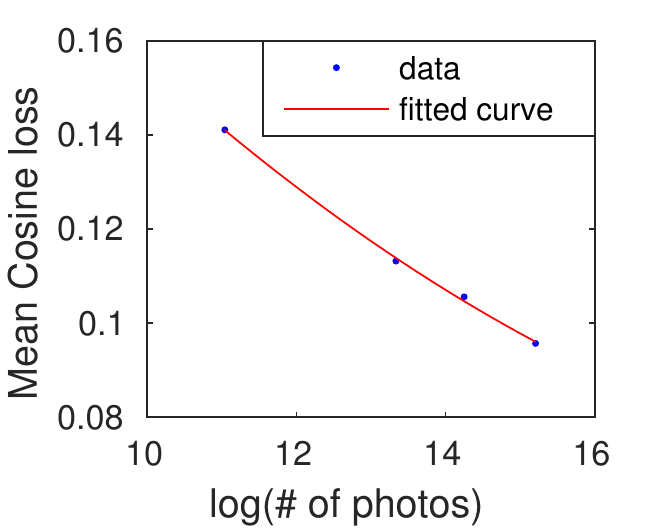}

  \end{subfigure}
  \begin{subfigure}{0.23\textwidth}
		\includegraphics[width=\textwidth]{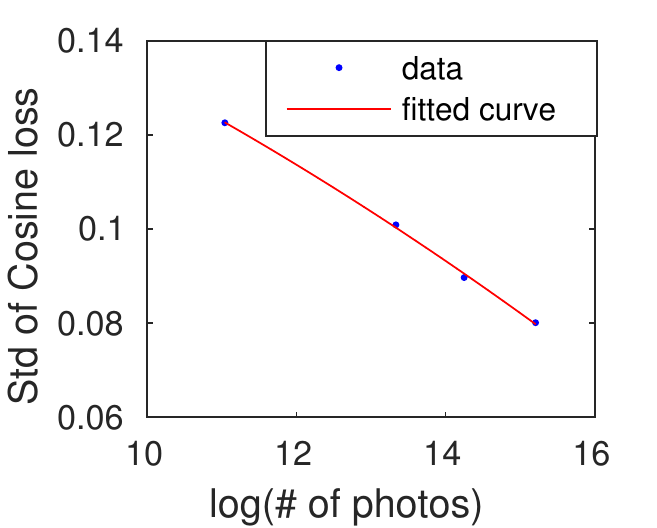}

  \end{subfigure}
	\caption{Cosines distance loss of models trained on different size of data.}
	\label{fig:cos_loss}
\end{figure}

To build $g(m_{\alpha}, m_{\beta})$ measuring the difference between two models trained on $m_{\alpha}$ images and $m_{\beta}$ images, we need to find a ``bridge'' to connect them. A nature idea is leveraging their loss that measures how far away they are from the perfect model and further implementing the triangle inequality to estimate the difference between themselves. While the classic \textit{softmax} loss is inappropriate here, cause it not satisfies the triangle inequality. Thus we used the Cosine distance again. We define the Cosine distance loss of a model $R(\cdot)$ as following:
\begin{equation}
\notag
\begin{array}{c@{\quad}l}
& \displaystyle{L_{cos}(R) = \sum_{a,b} l_{cos}(a,b,R)}\\
\mbox{where} & \displaystyle{
l_{cos}(a,b, R) = 
\begin{cases}
      1-|cos(R(a),R(b))|, \mbox{same identity;} \\
      |cos(R(a),R(b))|, \mbox{different identities.} \\
\end{cases}

}.
\end{array}
\end{equation}
And we count the mean and the standard deviation of Cosine distance loss for models trained on different data size. The results are showed on Fig~\ref{fig:cos_loss}.
Now, we can infer the mean ($\mu_{\beta}$) and the standard deviation ($\delta_{\beta}$) of the target model trained on $m_{\beta}$ photos, according to the fitted curve. Further, we assume the target model's and simulators' losses obey the normal distribution $\mathcal{N}(\mu_{\beta},\delta_{\beta}^2)$ and $\mathcal{N}(\mu_{\alpha},\delta_{\alpha}^2)$ respectively, and roughly estimate the difference by assuming it also obey the normal distribution $\mathcal{N}(\mu_{\beta}-\mu_{\alpha},\delta_{\beta}^2+\delta_{\alpha}^2)$. Thus, for a certain simulator $R(\cdot)$, we can calculate out what is the likelihood that $1-|cos(R(a),R(b))|$ plus the estimated difference will surpass $0.5$ for dodging attack or lower than $0.5$ for impersonation attack.

\begin{table}[tbh]
  \centering
    \caption{Statistics of $L_{cos}(R_{m_{\alpha}}) - L_{cos}(R_{m_{\beta}})$.}
	\begin{tabular}{|c|c|c|c|}
	\hline
	 $m_{\alpha}$ & $m_{\beta}$ & $\mu$ &  $\delta$\\
    \hline
	62258 & 1541811 & 0.0188 & 0.1176 \\
	\hline
	62258 & 4019407 & 0.0371 & 0.1225 \\
	\hline
	1541811 & 4019407 & 0.0163 & 0.1001 \\
	\hline
	\end{tabular}
	
	\label{tb:difference}
\end{table}
Besides, in Table~\ref{tb:difference}, we list true values of the distance of some pairs of $(m_{\alpha},m_{\beta})$, and observe that, along with the increase of data size, the simulator becomes similar to the target, and the impact of their data size difference becomes less prominent (small $\mu$ and $\delta$).

\section{Deduction of Equation 4}
\label{app:softmax}

For a DNN model, 
\begin{equation}
\notag
\begin{array}{c@{\quad}l}
 & Z_{(j)}(x) =  W_{(j)}^T R(x) + B_{(j)} 
\end{array}
\end{equation}
And for an ideally-trained DNN model $F^*$,
\begin{equation}
\notag
\begin{array}{c@{\quad}l}
 &  Z^*(x_a)_{(u)} = W_{(u)}^T R^*(x_a) =  0, \forall u \neq a \\
 & R^*(x_a) = t W_{(a)}, t \in \mathcal{R}
\end{array}
\end{equation}
as the perfection of the model. Thus, for a tuple $(x_a, x^{(\lambda)}, x_b)$,
\begin{equation}
\notag
\begin{array}{c@{\quad}l}
 F^*(x^{(\lambda)})_{(a)} &=  \frac{\exp(Z^*_{(a)})}{\exp(Z^*_{(a)}) + \exp(Z^*_{(b)})} \\
 &=  \frac{1}{1 + \exp(Z^*_{(b)}-Z^*_{(a)} )}
 
\end{array}
\end{equation}

Based on the NLL property, we know that $cos(R^*(x^{(\lambda)}), R^*(x_a)) = 1-\lambda$ and $cos(R^*(x^{(\lambda)}), R^*(x_b)) = \lambda$ hold true. So we get:
\begin{equation}
\notag
\begin{array}{c@{\quad}l}
 & R^*(x^{(\lambda)})^T R^*(x_a) = (1-\lambda) \|R^*(x_a)\|_2 \|R^*(x^{(\lambda)})\|_2 \\
 & R^*(x^{(\lambda)})^T R^*(x_b) = \lambda \|R^*(x_b)\|_2 \|R^*(x^{(\lambda)})\|_2 \\
\end{array}
\end{equation}
Using the results previously get we can also get:
\begin{equation}
\notag
\begin{array}{c@{\quad}l}
 & W(a)^T R^*(x^{(\lambda)}) = \zeta_a R^*(x_a)^T R^*(x^{(\lambda)}), \\
\end{array}
\end{equation}
where 
\begin{equation}
\notag
\begin{array}{c@{\quad}l}
 &\zeta_a = \|W(a)\|_2/\|R^*(x_a)\|_2 \\
\end{array}
\end{equation}
So,
\begin{equation}
\notag
\begin{array}{c@{\quad}l}
 & Z^*_{(a)}(x^{(\lambda)}) = (1-\lambda) \|W_{(a)}\|_2 \|R^*(x^{(\lambda)})\|_2
\end{array}
\end{equation}
Using the same, we get:
\begin{equation}
\notag
\begin{array}{c@{\quad}l}
 & Z^*_{(b)}(x^{(\lambda)}) = \lambda \|W_{(b)}\|_2 \|R^*(x^{(\lambda)})\|_2
\end{array}
\end{equation}
As the conclusion,
\begin{equation}
\notag
\begin{array}{c@{\quad}l}
 & Z^*_{(b)}(x^{(\lambda)}) - Z^*_{(a)}(x^{(\lambda)})  = \alpha \lambda - \beta \\
\end{array}
\end{equation}
 where 
\begin{equation}
\notag
\begin{array}{c@{\quad}l}
 & \alpha = \|R^*(x^{(\lambda)})\|_2 (\|W_{(b)}\|_2 + \|W_{(a)}\|_2) \\
 & \beta = \|R^*(x^{(\lambda)})\|_2 \|W_{(a)}\|_2 
\end{array}
\end{equation}
Besides, for a ideally-trained DNN model, we can assume that $F^*(x^{0.5})_(a) = 0.5$. From this assumption we can infer that $\alpha = 2\beta$, which is what we desired.

Combined, $F^*(x^{(\lambda)})_{(a)} = (1+\exp(2 \beta \lambda - \beta))^{-1}$

\section{Performance in Difference Distance}
\label{app:dist}

Without querying the target model, naturally the adversarial examples discovered by EXCIT, through simulating a ``better'' model, tend to be farther away from the subject. What we want to know, however, is for a given distance constraint, whether the examples found by our approach still have a higher probability of success, compared with the attack without the NLL enhancement. For this purpose, we need to modify the objective function of the DNN to limit its search within a given distance (in terms of $L_2$ distance) constraint, as follows:
\begin{equation}
\notag
\begin{array}{c@{\quad}l}
\mbox{minimize} & \displaystyle{ \exp (\| tanh(w)-x\|_2 - \gamma)+ f(w) }.
\end{array}
\end{equation}
Here we use an $exp$ function that penalizes the $L_2$ distance when exceeding $\gamma$. More specifically, we calculated its derivative as follows:
\begin{equation}
\notag
\exp(L_2 - \gamma) \cdot \frac{\tanh(w)-x}{\|\tanh(w)-x\|_2} \bigtriangledown_w \tanh(w)+ 1 \cdot \bigtriangledown_w f(w)
\end{equation}
where $L_2$ represents $\|\tanh(w)-x\|_2$. As we can see here, When $L_2 > \gamma$, the component involving the $exp$ function grows quickly, moving the objective function away from the optimality. Therefore, in the optimal situation, $L_2$ should not exceed $\gamma$ much.

%The derivative includes two terms: one with $\bigtriangledown_w \tanh(w)$ and the other involving $\bigtriangledown_w f(w)$. When $L_2 \ll \gamma$, the coefficient of the first term approaches $0$. When $L_2 \approx \gamma$, the first coefficient is comparable with $1$, the second coefficient. When $L_2 \gg \gamma$, the first coefficient is exponentially large. Thus, in the optimal situation, $L_2$ will not exceed $\gamma$ much.

\begin{figure}[ht]
	\centering
  \begin{subfigure}{0.15\textwidth}
		\includegraphics[width=\textwidth]{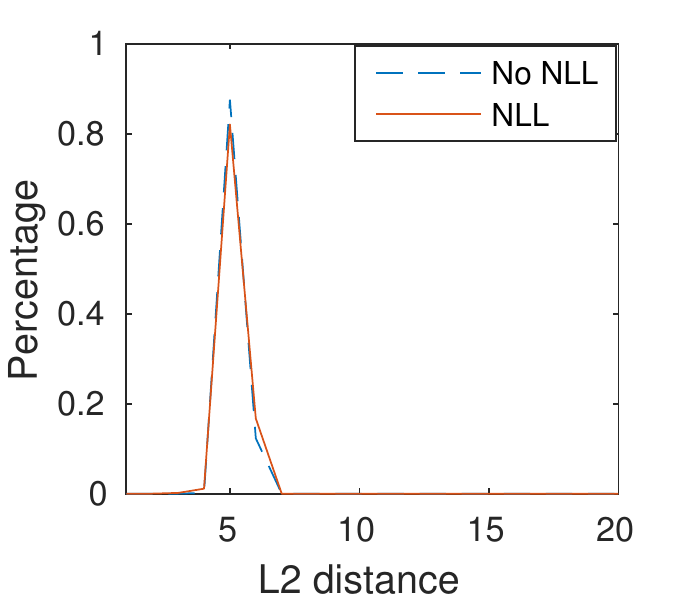}

  \end{subfigure}
  \begin{subfigure}{0.15\textwidth}
		\includegraphics[width=\textwidth]{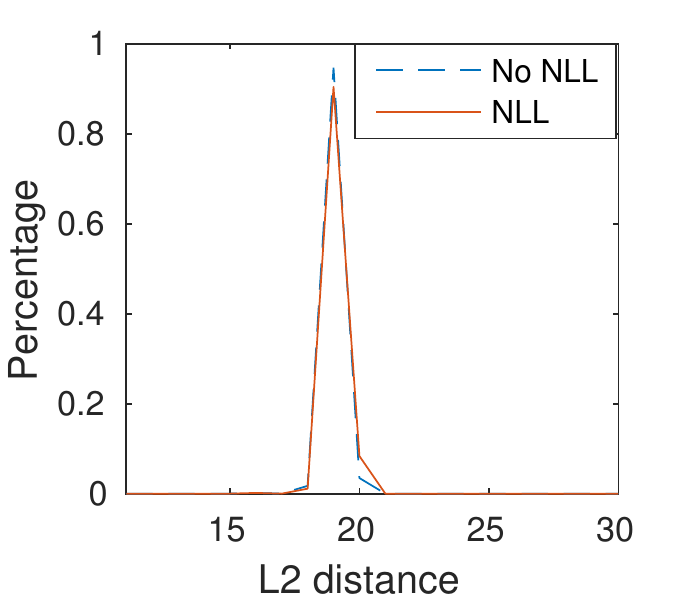}

  \end{subfigure}
  \begin{subfigure}{0.15\textwidth}
		\includegraphics[width=\textwidth]{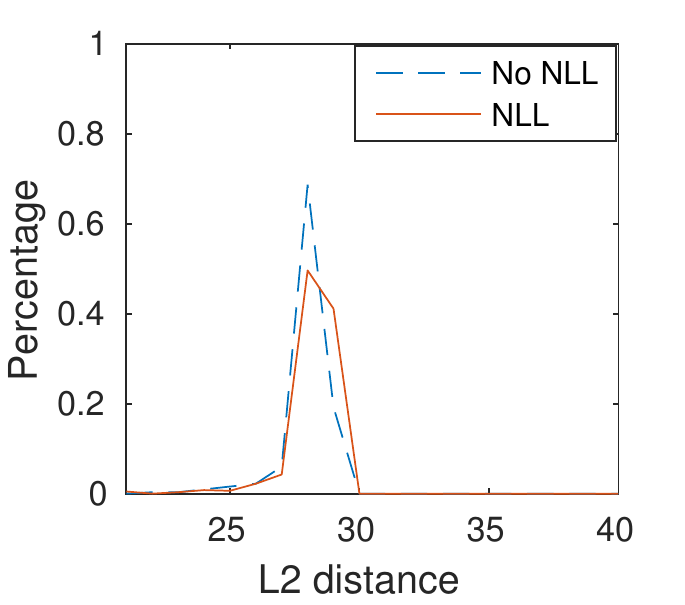}

  \end{subfigure}
	\caption{Distributions of modifications under different $\gamma$.}
	\label{fig:rst_pert}
\end{figure}

In our research, we evaluated our approach using the objective function when $\gamma = 5$, $\gamma = 20$ and $\gamma = 30$, and exploiting 4 EXCITs trained on 600K photos of 100K identities to attack the target trained on 1.9M photos of 300K identities. The results are presented in Fig~\ref{fig:rst_pert} and Table~\ref{tb:dist}. As we can see, under various distances, the adversarial examples found by our approach are always much more transferable than the one without the NLL enhancement. In the meantime, our approach tends to pick up the examples away from the subject, given the fact that the NLL property moves the decision boundary of the substitute model (with regard to the subject and the victim) closer to the ideal one, making it harder to find the adversarial examples close to the subject's image, though once such an image is found, it is more likely to lead to a successful attack. 

\begin{table}[tbh]
  \centering
    \caption{Impersonation transferability of NLL-enhanced approach under different distance constraints.}
	\begin{tabular}{|c|c|c|c|}
	\hline
	  $\gamma=$ & 5 & 20 & 30\\
    \hline
	with NLL & 6.3\% &  51.3\% & 74\% \\
	\hline
    without NLL & 3.8\% &  21.8\% & 49.5\% \\
	\hline
	\end{tabular}
	
	\label{tb:dist}
\end{table}

\section{Restrict the modification to certain region}
\label{app:eye}
\begin{figure}[ht]
	\centering
  \begin{subfigure}{0.15\textwidth}
		\includegraphics[width=\textwidth]{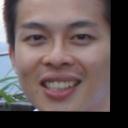}

  \end{subfigure}
  \begin{subfigure}{0.15\textwidth}
		\includegraphics[width=\textwidth]{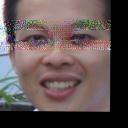}

  \end{subfigure}
  \begin{subfigure}{0.15\textwidth}
		\includegraphics[width=\textwidth]{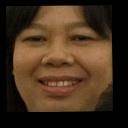}

  \end{subfigure}
  \begin{subfigure}{0.15\textwidth}
		\includegraphics[width=\textwidth]{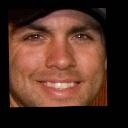}

  \end{subfigure}
  \begin{subfigure}{0.15\textwidth}
		\includegraphics[width=\textwidth]{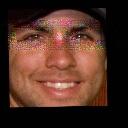}

  \end{subfigure}
  \begin{subfigure}{0.15\textwidth}
		\includegraphics[width=\textwidth]{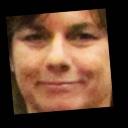}

  \end{subfigure}
	\caption{Successful impersonation attacks within restricted region.}
	\label{fig:imp_eye}
\end{figure}

The trivial method to restrict modifications within a certain region is to quench those derivatives out of the region, while finding the adversarial examples. However, in this setting, finding adversarial examples becomes harder than before. So it is need to totally release the constrain on the magnitude of modifications. We demonstrate two examples restricting modifications around eyes on Fig~\ref{fig:imp_eye}. We observe that the modifications become severe: the $L_2$ distance between generated adversarial example and the original photo of the first case is 24.63, and of the second case is 27.04.

\end{appendices}

\end{document}